\documentclass[letterpaper]{article}
\usepackage{uai2019}
\usepackage[margin=1in]{geometry}

\usepackage{times}

\def\fighome{./}
\def\bibhome{./}
\usepackage{times}
\usepackage{xcolor}
\usepackage{soul}
\usepackage[utf8]{inputenc}
\usepackage[small]{caption}

\usepackage{natbib}
\usepackage{paralist}
\usepackage{amsmath}
\usepackage{amsthm}
\usepackage{amssymb}
\usepackage{algorithm}
\usepackage[noend]{algorithmic}
\usepackage{graphicx}
\usepackage{pgfplots}
\usepackage{hyperref,url}
\usepackage{psfrag}

\def\Pbb{{\mathbb P}}
\newcommand{\llvert}{\left\vert\vphantom{\frac{1}{1}}\right.}

\newcommand\independent{\protect\mathpalette{\protect\independenT}{\perp}}

\newcommand{\Nb}{\text{nbd}}
\newcommand{\RG}{\text{RG}}
\newcommand{\Sg}{\text{Sg}}
\newcommand{\dist}{\text{dist}}

\newcommand{\Trip}{\text{Trip}}

\def\Rbb{\mathbb{R}}
\def\Ebb{\mathbb{E}}
\def\tha{{\mbox{\tiny th}}}
\def\independenT#1#2{\mathrel{\rlap{$#1#2$}\mkern2mu{#1#2}}}

\DeclareMathOperator{\Adj}{\mathcal{N}}
\DeclareMathOperator{\poly}{poly}
\def\calT{{\mathcal{ T}}}

\newtheorem{theorem}{Theorem}[section]
\newtheorem{lemma}[theorem]{Lemma}
\newtheorem{definition}[theorem]{Definition}
\newtheorem{assumption}[theorem]{Assumption}

\newtheorem{property}[theorem]{Property}
\floatname{algorithm}{Procedure}

\newcommand\mytitle{Guaranteed Scalable Learning of Latent Tree Models}

\title{\mytitle}

\author{ {\bf Furong Huang~\thanks{~\ Email: furongh@cs.umd.edu.}} \\
University of Maryland\\
\And
{\bf Niranjan Uma Naresh}  \\
Microsoft\\
\And
{\bf Ioakeim Perros}   \\
HEALTH[at]SCALE\\
\AND
{\bf Robert Chen} \\
Flatiron Health\\
\And
{\bf Jimeng Sun} \\
Georgia Institute of Technology\\
\And
{\bf Anima Anandkumar}\\
California Institute of Technology\\
}

\begin{document}

\maketitle

\begin{abstract}
We present an integrated approach for structure and parameter estimation in latent tree graphical models. Our overall approach follows a ``divide-and-conquer'' strategy that learns models  over small groups of variables and iteratively merges onto a global solution.   
The structure learning involves  combinatorial operations such as minimum spanning tree construction and local recursive grouping; the parameter learning is based on the method of moments and on tensor decompositions. 
Our method  is guaranteed to correctly recover the unknown tree structure and the model parameters with low sample complexity for the class of linear multivariate latent tree models which includes discrete and Gaussian distributions, and Gaussian mixtures. Our bulk asynchronous parallel algorithm is implemented in parallel and the parallel computation complexity increases only logarithmically with the number of variables and linearly with dimensionality of each variable. 
\end{abstract}
%!TEX root = 0_uai_latent_tree_main.tex
\section{INTRODUCTION}
%\fhcomment{(Motivation: Why Latent Tree Graphical Model is Useful?)}
Latent tree graphical models are a popular class of latent variable models, where a probability distribution involving observed and hidden variables are Markovian on a tree.
Since the structure of (observable and hidden) variable interactions is approximated as a tree, inference on latent trees can be carried out exactly through a simple belief propagation~\citep{pearl1988probabilistic}.
Therefore, latent tree graphical models present a good trade-off between model accuracy and computational complexity.
They are applicable in many domains~\citep{Durbin:book,choi2012context,choi2012context2,wang2013beyond}, where it is natural to expect hierarchical or
sequential relationships among the variables through a hidden-Markov model. 
%For instance,  latent tree models have been employed for phylogenetic reconstruction~\citep{Durbin:book}, object recognition~\citep{choi2012context,choi2012context2} and human pose estimation~\citep{wang2013beyond}. 

%\fhcomment{(Disadvantage of existing Latent Tree Learning: 1. Serial Learning of Structure; 2. Parameter Estimation Not Guaranteed)}
The task of learning a latent tree model consists of two parts: \emph{learning the tree structure} and \emph{learning the parameters of the tree}. We list the {challenges} in learning a latent tree model as follows:
\begin{compactenum}%[leftmargin=0]%[noitemsep,topsep=0pt,parsep=0pt,partopsep=0pt]
 \item \textbf{Challenge 1: Consistent structure learning.} The location and the number of latent variables are hidden and the marginalized graph over the observable variables no longer conforms to a tree structure.
 
\item \textbf{Challenge 2: Consistent parameter estimation.} Parameter estimation in latent tree model is typically carried out through Expectation Maximization (EM) or other local search heuristics~\citep{choi2011learning}.
These methods have no consistency guarantees, suffer from the problem of local optima and are not easily parallelizable.%, which can increase drastically, as the problem dimension increases.
%Moreover, these EM and likelihood-based methods are not easily parallelizable.

\item \textbf{Challenge 3: Computational complexity of structure learning.} Complexity of existing algorithms are typically polynomial with the number of variables $p$ (i.e., observed nodes) as discussed in~\cite{anandkumar2011spectral,choi2011learning}. These methods are sequential in nature and therefore are not scalable for large $p$.

\item \textbf{Challenge 4: Efficient structure and parameter estimation in parallel.} Existing methods treat structure learning and parameter estimation sequentially -- the parameter estimation can only be done after completion of structure learning. Therefore it is highly inefficient if the goal is to estimate only a small subset of variables/nodes.%: the tree structure is first estimated, and then the model parameters are fitted to the observed data.  In this paper, we present an integrated approach to structure and parameter estimation locally and simultaneously across groups. However, the challenge is to coordinate and align the decompositions over different groups of nodes, since the hidden labels can be permuted over the different decompositions.
\end{compactenum}

\cite{choi2011learning} addressed \textbf{challenge 1} using recursive grouping algorithms for Gaussian and discrete variables only. It remains unclear how to extend to variables in high-dimensions.  As for \textbf{challenge 2}, there is no work for guaranteed parameter estimation of latent tree models: although~\cite{anandkumar2012tensor} proposed tensor decomposition mechanisms for simple latent variable models such as multi-view and mixture of Gaussian models, extending those tensor decomposition mechanisms to hierarchical models such as latent trees are nontrivial, and involves alignment of locally estimated parameters.
No existing work addresses \textbf{challenge 3} or implements a structure learning of latent tree in time less than polynomial with $p$.   Lastly for \textbf{challenge 4}, there is no obvious way to (and no prior work did) directly parallelize these sequential methods without losing global consistency guarantees. 

We close the loop of consistent learning of latent tree model in high-dimensions via an integrated parallel approach to simultaneous structure and parameter estimation. Our method overcomes all above challenges.% which {prohibit} efficient or guaranteed learning. 

\textbf{Benefits of integrated structure and parameter estimation.}
The locally implemented and yet globally consistent simultaneous recovery of structure and parameter is the key to our algorithm's efficiency. Without this integration, parameter estimation has to wait until the completion of structure recovery, making the algorithm intrinsically sequential and thus less efficient. 
Another attractive feature of our method is that it is amenable for user interaction allowing the user to  provide feedback and change the course of various stages of the algorithm in a smooth manner: a quality not found in other graphical model learning methods.
More precisely, the user can select neighborhoods for adding hidden variables, using scores such as BIC.
This is suggested in~\cite{choi2011learning}, but the re-estimation of parameters through EM and sequential execution makes it expensive. In our approach, no re-estimation of parameters are needed since
the  structure   and parameter estimation under our framework go hand-in-hand, i.e., as the structure of the tree is obtained, parameters are dynamically estimated. %we also dynamically estimate the parameters.

\textbf{Summary of Contributions}

%\fhcomment{Contributions compared to [10]: our contributions are explicitly written in ``Summary of Contributions'' and the second paragraph above that in section 1: We present a new algorithm that improves computational complexity of learning latent tree from poly(p) in [10] to log(p) (with enough distributed computation nodes). We provide a consistent parameter estimation method, whereas EM used in [10] is not consistent. }
\textbf{Theoretical contributions.}
% First, our algorithm automatically learns the latent variables and their locations.
\textbf{(1)} Given enough computational resources, our method achieves consistent latent tree structure learning with $\log(p)$ computational complexity in a ``divide-and-conquer'' manner, improving the state-of-the-art $\text{poly}(p)$ complexity. We present a rigorous proof on the global consistency of the structure and parameter estimation under the ``divide-and-conquer'' framework. 
Our consistency guarantees are applicable to a broad class of linear multivariate latent tree models including discrete distributions, continuous multivariate distributions (e.g. Gaussian), and mixed distributions such as Gaussian mixtures.
This model class is much more general than discrete models prevalent in most previous works on latent tree models~\citep{mossel2005learning,mossel2007distorted,erdos1999few,anandkumar2013learning}. 
\textbf{(2)} Our algorithm guarantees consistent latent tree parameter estimation using inverse method of moments and tensor decomposition, the first guarantee for consistent parameter estimation in latent tree whose sample complexity is $\log(k)$. In contrast, the previous state-of-the-art is the EM algorithm~\citep{choi2011learning}. In addition, we extend tensor decomposition~\citep{anandkumar2012tensor} in models with simple structure to hierarchical tensor decomposition for more complex models. 
\textbf{(3)} Moreover, we carefully integrate structure learning with parameter estimation, based on tensor spectral decompositions~\citep{anandkumar2012tensor}.
The locally implemented and yet globally consistent recovery of structure and parameter simultaneously is the key for the efficiency of our algorithm. Without this integration, parameter estimation has to wait until the completion of structure recovery, making the algorithm intrinsically sequential and thus less efficient. 
%The parameter learning for any triplet of observed nodes on the latent tree can be carried out via standard tensor decomposition, as in~\cite{anandkumar2012tensor}.
Finally,  our approach  has a high degree of parallelism, and is {\em bulk asynchronous }parallel~\citep{gerbessiotis1994direct}. 
Thus, we propose a parallel and an integrated method for structure and parameter estimation without sacrificing on global correctness guarantees.

\textbf{Empirical justification.}We demonstrate that our algorithm is fast and scalable up to thousands of nodes and hundreds of thousands of data dimensions -- for example, it takes about 1 minute to run our method on nine nodes each with a dimensionality of 100,000, and about 70 minutes   when the number of nodes is 729 on a single workstation. Conceivably our method can be scalable to even larger dimensions by employing a cloud based implementation of the method. In our experiments,  we obtain a high level of accuracy, for both structure and parameter recovery, even as the problem dimensions increase.  In contrast, the EM method is stuck in local optima and has bad accuracy in parameter estimation, even for an extremely small example with nine nodes.   Thus, we demonstrate a scalable and guaranteed approach for learning latent tree graphical models.

%In addition to the aforementioned technical contributions, we demonstrate utility by applying it to two real datasets from the healthcare domain. 

\textbf{Application contribution.} In this work, we use latent tree model for discovering a hierarchy among diseases based on co-morbidities exhibited in patients' health records, i.e. co-occurrences of diseases in patients.  
In particular, two large healthcare datasets of 30K and 1.6M patients are used to build the latent disease trees. 
Our algorithm is used to discover hidden patterns, or concepts reflecting co-occurrences of particular diagnoses in patients in outpatient and intensive care settings. While such a task is currently done through manual analysis of the data, our method provides an automated way to discover novel clinical concepts from high dimensional, multi-modal data. Clinically meaningful disease clusters
are identified as shown in fig~\ref{Fig:tree_mimic2_1}.% and ~\ref{Fig:tree_mimic2_2}.%Quantitative measures such as Robinson Foulds metric are much better with our method compared to non-probabilistic methods such as agglomerative clustering.
%\fhcomment{Incorporate health care experiment contributions. }
%
%
%
%\fhcomment{Replace this example with the health care data. }
%For instance, for contextual object recognition, a latent tree model has been shown to be effective in   modeling the co-occurrence of objects in natural images~\citep{choi2012context2}, and the observed nodes in the latent tree correspond to the objects in images, and the hidden variables can be interpreted as latent contexts or categories.
%While~\citep{choi2012context2} employs   annotated text as observed variables, when such annotations  are unavailable, we can utilize visual features  of the objects (e.g. its response to object bank filters~\citep{li2010object} as observed data, and thus, we can have high dimensional (continuous) observed variables.

\begin{figure*}[!hbtp]
	%\centering
	\includegraphics[width=\textwidth]{\fighome/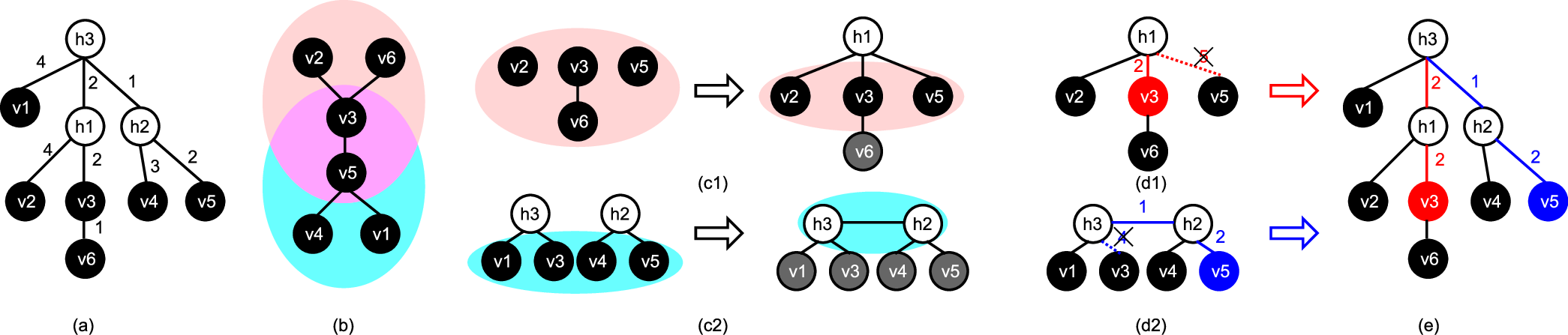}
	%\vspace{-0.6em}
	\caption{\small \textbf{(a)} Ground truth latent tree to be estimated, numbers on edges are \emph{multivariate information distances}. Filled nodes are observed and blank nodes are hidden. \textbf{(b)} MST constructed using the \emph{multivariate information distances}. $v_3$ and $v_5$ are internal nodes (leaders). Note that \emph{multivariate information distances} are additive on latent tree, not on MST. \textbf{(c1)} LRG on $\Nb[v_3,\text{MST}]$ to get local structure $\Adj_3$. Pink shadow denotes the active set. Local parameter estimation is carried out over triplets with joint node, such as ($v_2$, $v_3$, $v_5$) with joint node $h_1$. \textbf{(c2)} LRG on $\Nb[v_5,\text{MST}]$ to get local structure $\Adj_5$. Cyan shadow denotes the active set. \textbf{(d1)}\textbf{(d2)} Merging local sub-trees. Path($v_3$,$v_5$; $\Adj_3$) and path($v_3$,$v_5$; $\Adj_5$) conflict.  \textbf{(e)} Final recovery. }
	\label{Fig:StructureLearning}
\end{figure*}

%\begin{enumerate}
%\item Integrated structure and parameter estimation
%\item Divide and Conquer manner, parallel implementation
%\item Applied to health care datasets such as disease networks
%\end{enumerate}

%

%
%We apply our latent tree learning method to learn  hierarchies of words using co-occurrence data from a document corpus. Traditional topic modeling usually involves ``flat'' models such as the LDA. Recently, a hierarchical topic model is presented by \cite{kim2013variational}, but it assumes the hierarchy is   known and only learns the parameters. In contrast, our method learns both the structure and the parameters. Also, the work in \cite{nguyen2013lexical} is one of the several works similar to ours in the sense that they learn a latent hierarchy of topics without requiring a fixed structure. The main difference, however, is that they cannot have observed variables as internal nodes.

%!TEX root = 0_uai_latent_tree_main.tex

\section{LATENT TREE GRAPHICAL MODEL}
We denote $[n]:=\{ 1, \ldots, n \}$. Let $\mathcal{T}:= \left( \mathcal{V}, \mathcal{E}\right)$ denote the ground-truth undirected tree with  vertex set $\mathcal{V}$ and  edge set $\mathcal{E}$.
The \emph{neighborhood} of a node $v_i$ on tree $\mathcal{T}$, $\Nb[v_i,\mathcal{T}]$, is the set of nodes to which $v_i$ is directly connected on the tree.
Leaves which have a common neighboring node are defined as \emph{siblings}, and the common node is referred to as their {\em parent}. Let $N$ denote the number of samples. An example of latent tree is depicted in Figure~\ref{Fig:StructureLearning}(a).

There are two types of variables on the nodes, namely, the observable variables, denoted by $\mathcal{X} := \left\{x_1,\ldots,x_p\right\}$ ($p := \vert \mathcal{X} \vert$), and  hidden variables, denoted by $\mathcal{H}:=\left\{h_1,\ldots,h_m\right\}$ ($m := \vert \mathcal{H} \vert$).
Let $\mathcal{Y} := \mathcal{X} \cup \mathcal{H}$ denote the complete set of variables and let $y_i$ denote the random variable at node $v_i\in \mathcal{V}$, and similarly let $y_A$ denote the set of random variables in set $A$.
A \emph{graphical model} is defined as follows: given the neighborhood $\Nb[v_i,\mathcal{T}]$ of any node $v_i \in \mathcal{V}$, the variable $y_i$ is  conditionally independent  of the rest of the variables in $\mathcal{V}$, i.e., $y_i \independent y_j | y_{\Nb[v_i,\mathcal{T}]},\  \forall v_j\in \mathcal{V}\backslash \left\{v_i \cup \Nb[v_i,\mathcal{T}]
\right\}$.

\paragraph{Linear models.}We consider the class of linear latent tree models. %which includes discrete distributions, Gaussian multivariate models and Gaussian mixtures~\cite{anandkumar2011spectral}.  
The observed variables $x_i$ are random vectors of length $d_i$,
%\jscomment{all observed variables have to have the same dimensionality $d$? Is this easy to generalize to different dimensionalities for different variables? If easy, we should state this is a simple extension. If it is really easy, we should say how exactly to do that zero padding?}
 i.e., $x_i \in \Rbb^{d_i},\  \forall i\in [p]$ while the latent nodes  are $k$-state categorical variables, i.e., $h_i \in \{e_1, \ldots, e_k \}$, where $e_j\in \Rbb^k$ is the $j^{\tha}$ standard basis vector. Although $d_i$ can vary across variables, we use $d$ for notation simplicity. In other words, for notation simplicity, $x_i \in \Rbb^{d},\  \forall i\in [p]$ is equivalent to $x_i \in \Rbb^{d_i},\  \forall i\in [p]$. 
For any variable $y_i$ with neighboring hidden variable $ h_j$, we assume a linear relationship: 
$ \Ebb[y_i |h_j] = A_{y_i\llvert h_j}  h_j,$ 
where transition matrix $A_{y_i\llvert h_j} \in \mathbb{R}^{d\times k}$ is assumed to have full column rank, $\forall y_i,h_j\in \mathcal{V}$. 
This implies that $k\leq d$, which is natural if we want to enforce a parsimonious model for fitting the observed data. If two observable variables interact through at least a hidden variable (i.e., there is at least a hidden variable along the path between the two nodes), we have $\mathbb{E}[y_a y_b^\top] =  \sum\limits_{e_i}\mathbb{E}[h_j=e_i]  A_{y_a\llvert h_j=e_i} A_{y_b\llvert h_j=e_i}^\top.$ 
We see that $\mathbb{E}[y_a y_b^\top]$ is of rank $k$ since $A_{y_a\llvert h_j=e_i}$ or $A_{y_b\llvert h_j=e_i}$ is of rank $k$.
We consider the class of tree models where it is possible to recover the latent tree model uniquely. 

\begin{assumption}[Structure Identifiable Condition~\citep{choi2011learning}]
Each hidden variable has at least three neighbors (which can be either hidden or observed). 
\end{assumption}
\emph{Remark:} Our structure identifiable condition ensures a minimal latent tree with no redundant nodes. Therefore our goal is to consistently learn the set of identifiable latent tree model. 

\begin{assumption}[Parameter Identifiable Condition]
(1) The pairwise correlation matrix $\mathbb{E}\left[x_a x_b^\top \right]$,  between neighboring observable variables  $x_a$ and $x_b$,  is of rank $k$.  (2) Any two variables connected by an edge in the tree model are neither perfectly dependent nor independent. 
\end{assumption}
\emph{Remark:} (1) When two observed nodes are directly connected according to the structure learned, the conditional probability decomposes into $k$ factors. This assumption is mild and applies to various applications where observed nodes have intrinsic memberships (low dimensional representation) through which they interact with other nodes.  (2) This condition is necessary for the identifiability of parameters as stated in~\cite{choi2011learning}. 

% For a pair of (observed or hidden)  variables $y_a$ and $y_b$, consider the \emph{pairwise correlation matrix}  $\mathbb{E}\left[y_a y_b^\top \right]$  where the expectation is over samples.

\textbf{Learning objective.} Our  {goal} is to learn the ground-truth structure $\mathcal{T}:= \left( \mathcal{V}, \mathcal{E}\right)$ and parameter $A_{y_i\llvert h_j}$ given $N$ examples of $\left\{\left(x_1^{(j)},\ldots,x_p^{(j)}\right)\right\}_{j=1}^N$.

%\paragraph{Minimal latent tree?.}
%\fhcomment{\textbf{Clarification on why assuming ``two observable variables interact through at least a hidden variable''}: This assumption is due to the linear model considered in this paper (1) For identifiability of parameter estimation of latent trees, we assume a linear generative model (in section 2) where each observed variable is generated through a hidden variable through linear transformation, i.e., conditional probability of observed variable given hidden variable P(x|h) ). (2) Therefore linear model implies that two observed variables interact through at least a hidden variable. When two observed nodes are directly connected according to the structure learned, the parameter estimation factorize the conditional probability, which is implicitly introducing a hidden variable for parameter estimation. (3) In practice, linear models are applied successfully to a wide class of latent variable models such as mixture of Gaussian and latent dirichlet allocation. }

%!TEX root = 0_uai_latent_tree_main.tex

\section{APPROACH OVERVIEW}\label{sec:overview}
\vspace{-0.7em}
The overall approach is depicted in Figure~\ref{Fig:StructureLearning}, where (a) and (b) show the data preprocessing step,  
 (c) - (e) illustrate the divide-and-conquer step for structure and parameter learning. We will describe each step in details in the later sections.

We start with the parallel computation of 
pairwise \emph{multivariate information distances} (defined in Definition~\ref{def:info_dist}) between all pairs of observed variables. Information
distance roughly measures the correlation between different pairs of observed variables and requires SVD computations. For this example in Figure~\ref{Fig:StructureLearning}(a), the multivariate information distances between pairs of $\{v_1,\ldots,v_6\}$ will be estimated empirically via Equation~\eqref{eqn:info_dist_est}. Note that the tree structure is hidden and unknown.

Then, as depicted in Figure~\ref{Fig:StructureLearning}(b), a Minimum Spanning Tree (MST) over the estimated pairwise multivariate information distances is constructed over observable variables in parallel~\citep{bader2006fast}. The local groups (pink and cyan) are also obtained through MST so that they are available for the structure and parameter learning step that follows.

The structure and parameter learning is done jointly through a divide-and-conquer strategy. 
Figure~\ref{Fig:StructureLearning}(c) illustrates the divide step (or local learning), where local structure (LRG) and parameter estimation (tensor decomposition) is performed (Procedure~\ref{algo:plrg}). 
%This results in huge computational savings. Our learning procedure consists of  localized combinatorial computations (local recursive grouping) for structure learning along with  tensor decompositions for  parameter estimation. 
%No coordination is needed across different groups during this process.

Our algorithm also performs the local merge to obtain group level structure and parameter estimates. 
As shown in Figure~\ref{Fig:StructureLearning}(d1) and (d2), after the local structure and parameter learning is finished within the groups,
 we perform merge operations among groups, again guided by the Minimum Spanning Tree structure. For the structure estimation it consists of  a  union operation of sub-trees (Procedure~\ref{algo:pmac});  for the parameter estimation, it consists of  linear algebraic operations (Procedure~\ref{algo:alignment}) -- Since our method is unsupervised, an alignment procedure of the hidden states is carried out which finalizes the global estimates of the tree structure and the parameters.

% A key feature is that the size of these groups are typically  small: we prove that the size is bounded by a function involving the maximum degree of the unknown latent tree and   the extent to which  the model parameters are ``homogeneous''. If the latent tree degree and depth are constant (e.g. the hidden Markov model (HMM)) and the parameters are not too ``heterogeneous'',  we obtain constant sized groups, and thus, a high level of parallelization.
%The key point is that we retain global consistency guarantees, despite parallelization, since we carefully selects the groups over which the sub-trees are constructed. 

%!TEX root = 0_uai_latent_tree_main.tex
\vspace{-0.5em}
\section{STRUCTURE LEARNING}\label{sec:structure}
\vspace{-0.7em}
%The information distance was first introduced and its additivity proven in~\cite{lake1994reconstructing}.
Structure learning in graphical models involves finding the underlying Markov graph, given the observed samples. For latent tree models, structure can be estimated via distance based methods. This involves computing certain {\em information} distances between any pair of observed variables,  and then finding a tree which fits the computed distances. 
%\fhcomment{comment out this subtitle. Start from the MST construction, and then in appendix say that the distance calculation is crucial. }

\textbf{Multivariate information distances: }
%\fhcomment{shorten this information distance section. }
We propose a distance metric, that is additive on the ground truth tree, for multivariate linear latent tree models. 
For a pair of (observed or hidden)  variables $y_a$ and $y_b$, consider the pairwise correlation matrix  $\mathbb{E}\left[y_a y_b^\top \right]$ (the expectation is over samples). %\jscomment{why is rank $k$? not clear to me what the expectation is over. this is a fundamental definition which should be formally introduced, maybe in the section 2}.\fhcomment{Defined it in section 2 and discussed why it is of rank k.}  
Note that its rank is $k$, dimension of the hidden variables. %Below, we consider an information distance based on its rank-$k$ SVD.
\begin{definition}
\label{def:info_dist}
The multivariate information distance between nodes $i$ and $j$ is defined as
\begin{equation}
\label{eqn:info_dist}
\dist(v_a,v_b) := -\log \frac{\prod\limits_{i=1}^{k}\sigma_i\left(\mathbb{E}(y_a y_b^\top)\right)}{\sqrt{\det(\mathbb{E}(y_a y_a^\top)) \det(\mathbb{E}(y_b y_b^\top))}}
\end{equation}
where
$\{\sigma_1(\cdot),\ldots,\sigma_k(\cdot)\} $ are the top $k$ singular values.
\end{definition}
\emph{Remark:} This is an extension of the distance measure to multivariate variables, which is not introduced in~\cite{choi2011learning}.
Note that definition~\ref{def:info_dist} suggests that this multivariate information distance allows heterogeneous settings where the dimensions of $y_a$ and $y_b$ are different (and $\geq k$). For finite number ($N$) of samples the empirical estimation of the multivariate information distance is estimated as
{\small{
\begin{align}
\label{eqn:info_dist_est}
\widehat{\dist}(v_a,v_b) =
 -\log \frac{\prod\limits_{i=1}^{k}\sigma_i\left(\sum\limits_{j=1}^N(y_a^{(j)} (y_b^{(j)})^\top)\right)}{\sqrt{\det(\sum\limits_{j=1}^N(y_a^{(j)} (y_a^{(j)})^\top)) \det(\sum\limits_{j=1}^N(y_b^{(j)} (y_b^{(j)})^\top))}}
\end{align}
}}
For latent tree models, we can find information distances which are provably {\em additive} on the underlying tree  in expectation, i.e.  the expected distance between any two nodes in the tree is the sum of distances along the path between them. %For instance, for the special case when the variables on the tree are (scalar) Gaussian, the negative logarithm of the correlation coefficient  corresponds to an additive information distance.

%\fhcomment{Clarification on the assumption of the ground truth latent tree structure and uniqueness of the tree: The latent tree structure we estimate is the minimal tree such that each latent variable has at least 3 neighbors, as it is the class of tree models where it is possible to recover the latent tree model uniquely. We explained it in details in Definition H.1. We will emphasize this condition in lemma 4.2 and in the main body in the revised paper.}

\begin{lemma}\label{lem:additive}
The multivariate information distance is additive on the tree $\mathcal{T}$, i.e., $\dist(v_a,v_c) = \dist(v_a,v_b) + \dist(v_b,v_c)$, where $v_b$ is a node in the path from $v_a$ to $v_c$ and $v_a$,$v_b$,$v_c\in \mathcal{V}$.
\end{lemma}
Refer to \href{https://drive.google.com/file/d/1ZkKVJjLrI1aMf-VevMJT1B1saXhtXcmK/view?usp=sharing}{Appendix}~\ref{apdx:additive} for proof. The empirical distances can be computed via rank-$k$ SVD of the empirical pairwise moment matrix $\hat{\Ebb}[y_a y_b^\top]$. %\jscomment{what is the exact definition? this is related to the early comment on a similar definition}. 
Note that the distances for all the pairs can be computed in parallel. %Moreover, we will see that the computed SVDs are also useful for parameter estimation through tensor decomposition, and we can store them for later use. Thus, by coordinating structure and parameter estimation we can save on computational costs.

To estimate the structure,  we could do reverse engineering to figure out where to introduce hidden nodes or how nodes are connected since the multivariate information distances should be additive on the ground-truth tree. Below we extend LRG in~\cite{choi2011learning} to be in parallel to consistently and efficieintly estimate the structure. 

\textbf{Formation of local groups via MST: }Once the empirical distances are computed, we construct a    Minimum Spanning Tree (MST), based on those distances. Note that the MST can be computed efficiently in parallel~\citep{vineet2009fast,website:Boruvka}. We now form groups of observed variables over which we carry out learning independently, without any coordination. These groups are obtained by the (closed) neigborhoods in the MST, i.e. an internal node and its one-hop neighbors form a group. The corresponding internal node is referred to as the {\em group leader}. See Figure~\ref{Fig:StructureLearning}(b).

\textbf{Local recursive grouping (LRG): }Once the groups are constructed via neighborhoods of MST,  we construct a sub-tree with hidden variables in each group (in parallel) using the recursive grouping introduced in~\cite{choi2011learning}, depicted in \ref{Fig:StructureLearning}(c1) and (c2).    
%\fhcomment{We refer to it as  the \emph{ local recursive grouping test}, and provide it in Procedure~\ref{algo:lrg}.}
The recursive grouping uses the multivariate information distances and decides the locations and numbers of hidden nodes. 
It proceeds by deciding which nodes are ``siblings'' or ``parent and child'', using the following property that determines a pair of siblings or a parent and child pair.

\begin{property}[Siblings]
	Define a potential function as $\Phi(v_i,v_j;v_a):=\dist(v_i,v_a)- \dist(v_j,v_a)$. A pair of observed nodes ($v_i$, $v_j$) are siblings with parent $v_l$, if the potential function is fixed $\forall v_a,v_b$ in the active set 
	\begin{equation*}
	\Phi(v_i,v_j;v_a) \equiv \Phi(v_i,v_j;v_b) = \dist(v_i,v_l) - \dist(v_j,v_l).
	\end{equation*}
\end{property}
%	 which proceeds as follows: consider two observed nodes $v_i,v_j$ which are siblings on the tree with a common parent $v_l$, and consider any other observed node $v_a$. 
From additivity of the (expected) information distances, we have $\dist(v_i,v_a)= \dist(v_i,v_l) + \dist(v_l,v_a)$ and similarly for $\dist(v_j,v_a)$. 
Thus, we have $\Phi(v_i,v_j;v_a):=\dist(v_i,v_a)- \dist(v_j,v_a)= \dist(v_i,v_l) - \dist(v_j,v_l)$, which is independent of node $v_a$. 

\begin{property}[Parent-Child]
A pair of observed nodes ($v_l$, $v_i$) is a parent ($v_l$) and child ($v_i$) pair , if $\forall v_a,v_b$ in the active set 
\begin{equation*}
\Phi(v_i,v_l;v_a) = \dist(v_i,v_a)- \dist(v_l,v_a) = \dist(v_i,v_l)
\end{equation*}
\end{property}

Thus, comparing the quantity $\Phi(v_i,v_j;v_a)$ for all nodes $v_a$ allows us to determine whether $v_i$ and $v_j$ are siblings or parent-child. 
For instance, in (c1), firstly the active set is $\{v_2,v_3,v_5,v_6\}$,  $v_3$ and $v_6$ are detected as parent and child because for all other nodes in the active set, i.e., $v_2$ and $v_5$, we have $\Phi(v_6,v_3;v_2) = \dist(v_6,v_2) - \dist(v_3,v_2) = \Phi(v_6,v_3;v_5) = \dist(v_6,v_5)- \dist(v_3,v_5) = \dist(v_3,v_6)$; secondly the active set is updated to $\{ v_2,v_3,v_5\}$ (children are deleted), $v_2$ and $v_3$ are detected as siblings because for all other nodes in the active set, i.e., $v_5$, we have $\Phi(v_2,v_3;v_5) \equiv \Phi(v_2,v_3;v_5)$. Similarly $v_2$ and $v_5$ are detected as siblings and therefore   $v_2,v_3,v_5$ are detected as siblings in the second step.

Once the siblings are inferred, the hidden nodes are introduced, and the same procedure repeats to construct the higher layers. Note that whenever we introduce a new hidden node $h_{\text{new}}$ as a parent, we need to estimate multivariate information distance between $h_{\text{new}}$ and nodes in active set $\Omega$. This is discussed in~\cite{choi2011learning} in detail.

%\fhcomment{Our main contribution is not introduction of LRG but a new local/parallel framework with global consistency guarantee, therefore we only briefly introduce of the LRG method. We will provide more details in the final version.}

\paragraph{Finite samples}  Our algorithm produces consistent results even when there is finite number of samples and the estimation of $\dist(v_a,v_b)$ is noisy (see Equation~\eqref{eqn:info_dist_est}).

\emph{Parent-child:} In the noiseless case, 	if {$\Phi (v_a, v_b; v_c) = \dist(v_a,v_b), \; \forall$ $v_c \in \Omega \backslash \{v_a, v_b\}$}, then $v_a$ is a leaf node and $v_b$ is its parent. 
However, in the noisy case, we modify it to 
if {$ |\widehat{\Phi}(v_a, v_b; v_c) - \widehat{\dist}(v_a,v_b)| \le \epsilon, \; \forall$ $v_c \in \Omega \backslash \{v_a, v_b\}$}, then $v_a$ is a leaf node and $v_b$ is its parent.

\emph{Siblings:} In the noiseless case,  if {$-\dist(v_a,v_b) < \Phi (v_a, v_b; v_c) = \Phi (v_a, v_b; v_c^\prime) < \dist(v_a, v_b),\forall v_c, v_c^\prime \in \Omega \backslash \{v_a, v_b\}$}, then $v_a$ and $v_b$ are siblings.
However, in the noisy case, we modify it to 
if  {$|\widehat{\Phi} (v_a, v_b; v_c) - \widehat{\Phi} (v_a, v_b; v_c^\prime)|\le \epsilon$, and $| \widehat{\Phi} (v_a, v_b; v_c)| <\widehat{\dist}(v_a,v_b)+\epsilon$, $\forall v_c, v_c^\prime \in \Omega \backslash \{v_a, v_b\}$}, then $v_a$ and $v_b$ are siblings.

The threshold in the procedure is not a simple heuristic but is supported by theory: Lemma 7.2 provides a consistency  guarantee on the learning of latent tree structure using the noisy pairwise distance. For a given precision $\epsilon$ and a given number of variables $p$, we derive the number of samples $N$ needed for our algorithm to be consistent.

We describe the LRG in details with integrated parameters estimation in Procedure~\ref{algo:plrg} in Section~\ref{sec:merging}. 
In the end, we obtain a sub-tree over the local group of variables. After this \emph{local recursive grouping test}, we store the neighborhood relationship for the leader $v_i$ using an adjacency list $\Adj_i$. We call the resultant local structure the \emph{latent sub-tree}.
 %This is different from~\cite{choi2011learning} since they serially introduce hidden nodes and form the global tree whereas we propose the divide-and-conquer approach which is amendable for parallel implementation. 

%!TEX root = 0_uai_latent_tree_main.tex

\section{PARAMETER ESTIMATION}\label{sec:parameter}
Along with the structure learning, we use a moment-based spectral learning technique for parameter estimation. This is a guaranteed and fast approach to recover parameters via moment matching for third order moments of the observed data. In contrast, traditional approaches such as Expectation Maximization (EM) suffer from spurious local optima and cannot provably recover the parameters.

\textbf{A latent tree with three leaves:} We first consider an example of three observable leaves $x_1,x_2, x_3$ (i.e., a triplet) with a common hidden parent $h$. We then clarify how this can be generalized to learn the parameters of the latent tree model.
Let $\otimes$ denote for the tensor  product. For example, if $x_1, x_2, x_3 \in \mathbb{R}^{d}$, we have $x_1 \otimes x_2\otimes x_3 \in \mathbb{R}^{d \times d\times d}$ and $[x_1 \otimes x_2\otimes x_3]_{ijk}=x_1(i)x_2(j)x_3(k)$.

\begin{figure}
\psfrag{Hidden}[]{$h$}
\psfrag{x1}[]{$x_1$}
\psfrag{x2}[]{$x_2$}
\psfrag{x3}[]{$x_3$}
\includegraphics[width=0.5\textwidth]{\fighome/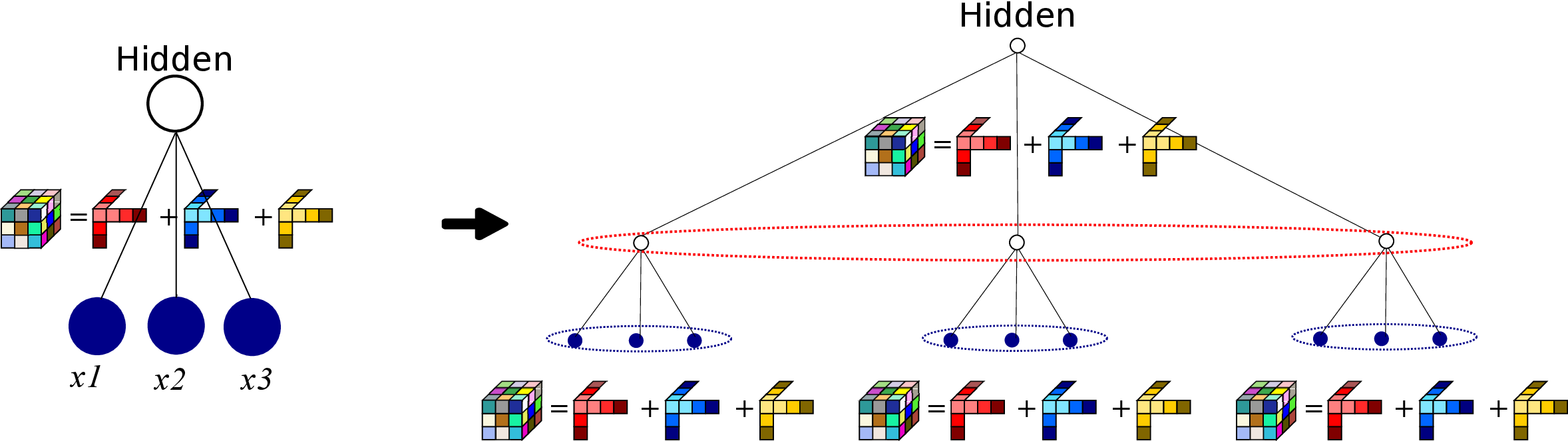}
\caption{Parameter estimation in latent tree using hierarchical tensor decomposition.}\label{fig:flat2hier}
\end{figure}

% Question on why the expectation only depends on one hidden variable:  The hidden variable separates the two observed variables on the latent tree, therefore the two observed variables are independent given the hidden variable according to the definition of graphical model.

\begin{property}[Tensor decomposition for triplets]  For a linear latent tree model with three observed nodes $v_1,v_2,v_3$ with a common hidden parent $h$, we have
%\begin{equation}
$\mathbb{E}(x_1\otimes x_2 \otimes x_3) = \sum_{r=1}^{k} \Pbb[h=e_r] A_{x_1|h}^r \otimes A_{x_2|h}^r \otimes A_{x_3|h}^r$, 
%\end{equation}
where $A_{x_i|h}^r =\mathbb{E}(x_i|h=e_r)$, i.e., $r^{\text{th}}$ column of the transition matrices from $h$ to $x_i$. The hidden variable separates the observed variables on the latent tree, therefore the observed variables are conditionally independent given the hidden variable, according to the definition of graphical model.
The tensor decomposition method of~\cite{anandkumar2012tensor} provably recovers the parameters $A_{x_i|h}$, $\forall i\in [3]$, and $\Pbb[h]$.\end{property}\label{lem:para_est_def}

\emph{Remark:} To recover the model parameter $A_{x_i|h}$, we need an empirical estimation of $\mathbb{E}(x_1\otimes x_2 \otimes x_3)$, computed as $\widehat{\mathbb{E}}(x_1\otimes x_2 \otimes x_3) = \frac{1}{N}\sum\limits_{j=1}^N x_1^{(j)} \otimes x_2^{(j)}  \otimes x_3^{(j)}$. We then use tensor decomposition on $\widehat{\mathbb{E}}(x_1\otimes x_2 \otimes x_3)$ to estimate $A_{x_i|h}$. Once we obtain the conditional distribution of observed variable given the hidden variable $\Pbb(x_i|h)$ and the marginal distribution of the $\Pbb(h)$ after tensor decomposition, we sample from the posterior distribution $\Pbb(h|X)$ for each corresponding observed example in parameter estimation procedure.

\textbf{Tensor decomposition for learning latent tree models: }We employ the above approach for learning latent tree model parameters as follows: for every triplet of variables $y_a$, $y_b$, and $y_c$ (hidden or observed), we consider the hidden variable $h_i$ which is the \emph{joining point} of $y_a,y_b$ and $y_c$ on the tree, i.e., the node that path$(y_a,y_b)$, path$(y_b,y_c)$ and path$(y_a,y_c)$ go through. They form a {\em triplet} model, for which we employ the tensor decomposition procedure.   However, it is wasteful to do it over all the triplets in the latent tree.  %Moreover, triplets of nodes which are far away in the tree tend to have low correlation, and thus, require a large number of samples to estimate consistently.
In the next section, we demonstrate how we efficiently estimate the parameters simultaneously as we learn the structure, and minimize the tensor decompositions required for estimation.

The samples from the posterior distribution $\Pbb(h|X)$ are required for parameter estimation (tensor decomposition), although it is not required for the computation of additive distance. The tensor decomposition for each triplets on the tree requires sample estimation of moments on variables. When estimating parameters related to the observed variables, no sampling is needed as samples are observed. However estimating parameters for the internal hidden nodes requires sampling from the hidden variables to form the moments of the triplets.

%Issues such as alignment of hidden labels across different decompositions will also be addressed.
%\paragraph{Hidden Node Estimation via Least Squares Approximation:} We approximate the posterior over a hidden node using a least squares approximation. Consider a hidden node $h$ with children $x_i, \forall i \in [l]$ with transition matrix from $h$ to $x_i$ denoted by $\zeta_i$.
% We have
%$\mathbb{E}[x_i] = \zeta_i \mathbb{E}[h] + \epsilon_i.
%$
%Now, we concatenate the samples $x_i$ to obtain $C\in \mathbb{R}^{l\times d}$ and concatenate the transition matrices $\zeta_i$ to obtain a $\zeta \in \mathbb{R}^{ld \times k}$. Hence, we infer the hidden states by $\mathbb{E}[h] = \zeta^\dag C.$

%\fhcomment{Question on why the expectation only depends on one hidden variable:  The hidden variable separates the two observed variables on the latent tree, therefore the two observed variables are independent given the hidden variable according to the definition of graphical model.}

%!TEX root = 0_uai_latent_tree_main.tex
\vspace{-0.5em}
\section{INTEGRATED %SIMULTANEOUS 
STRUCTURE AND PARAMETER ESTIMATION}
\label{sec:merging}
So far, we have described, in a high-level, procedures of structure estimation through local recursive grouping (LRG) and parameter estimation through tensor decomposition over triplets of variables, respectively. We now introduce an integrated and efficient approach which brings all these ingredients together. We provide merging steps to obtain a global model that is consistent, using the sub-trees and parameters learned over local groups.
\subsection{{Structure with Parameter Estimation}}
Intuitively, we find efficient groups of triplets to carry out tensor decomposition simultaneously, as we estimate the structure through recursive grouping. In LRG, pairs of nodes are recursively grouped as siblings or as parent-child. As this process continues, we carry out tensor decompositions whenever there are siblings presented as triplets. If there are only a pair of siblings, we find an observed node with closest distance to the pair. Once the tensor decompositions are carried out on the observed nodes, we proceed to structure and parameter estimation of the  added  hidden variables. The samples of the hidden variables can be obtained via the posterior distribution, which is learnt earlier through tensor decomposition. This allows us to predict information distances and third order moments among the hidden variables as process continues. See algorithm in Procedure~\ref{algo:plrg}. 
%{\LinesNumberedHidden
%\jscomment{check the algorithm, how come there are 2 input lines and line 1 of the algorithm is before input and output. output is 1 sub-tree, or many subtrees, if it is one, let's make that clear. also does that mean we need to call this procedure many times for each $v_i$? for loop is confusing, i suggest u remove that, and just say that in the text u need to call this for each $v_i$.}
 \begin{algorithm}[h]
 \caption{LRG with Parameter Estimation}
 \label{algo:plrg}
 \begin{algorithmic}[1]
% \REQUIRE Internal nodes $\mathcal{X}_\text{int}$ on MST.
% \FORALL {}
 \REQUIRE Internal nodes $\mathcal{X}_\text{int}$ on MST, for each $v_i \in \mathcal{X}_\text{int}$, active set $\Omega := \Nb[v_i;\text{MST}]$, precision $\epsilon$
  \ENSURE for each $v_i \in \mathcal{X}_\text{int}$, local sub-tree adjacency matrix $\Adj_i$, and $\widehat{\mathbb{E}}[y_a| y_b]$ for all $(v_a,v_b)\in \Adj_i$.
 	\STATE Active set $\Omega \leftarrow \Nb[v_i;\text{MST}]$
 	\WHILE{$\lvert \Omega\rvert> 2$}
 		\FORALL {$v_a,v_b\in \Omega$}
 		\IF %{$\Phi (v_a, v_b; v_c) = \dist(v_a,v_b), \; \forall$ $v_c \in \Omega \backslash \{v_a, v_b\}$}
 		{$ |\widehat{\Phi}(v_a, v_b; v_c) - \widehat{\dist}(v_a,v_b)| \le \epsilon, \; \forall$ $v_c \in \Omega \backslash \{v_a, v_b\}$}
 		\STATE $v_a$ is a leaf node and $v_b$ is its parent,
 		\STATE Eliminate $v_a$ from $\Omega$.
 		\ENDIF
 		\IF%{$-\dist(v_a,v_b) < \Phi (v_a, v_b; v_c) = \Phi (v_a, v_b; v_c^\prime) < \dist(v_a, v_b),\forall v_c, v_c^\prime \in \Omega \backslash \{v_a, v_b\}$}
 		 {$|\widehat{\Phi} (v_a, v_b; v_c) - \widehat{\Phi} (v_a, v_b; v_c^\prime)|\le \epsilon$, and $| \widehat{\Phi} (v_a, v_b; v_c)| <\widehat{\dist}(v_a,v_b)+\epsilon$, $\forall v_c, v_c^\prime \in \Omega \backslash \{v_a, v_b\}$}
 		\STATE $v_a$ and $v_b$ are siblings,eliminate $v_a$ and $v_b$ from $\Omega$, add $h_{\text{new}}$ to $\Omega$.
 		\STATE Introduce new hidden node $h_{\text{new}}$ as parent of $v_a$ and $v_b$.
 		\IF {more than 3 siblings under $h_{\text{new}}$}
 		\STATE find $v_c$ in siblings,
 		\ELSE
 		 \STATE find $v_c = \arg \min_{v_c\in \Omega} \dist(v_a,v_c)$.
 		\ENDIF
 		\STATE Estimate empirical third order moments $\widehat{\mathbb{E}}(y_a\otimes y_b\otimes y_c)$
 		\STATE Decompose $\widehat{\mathbb{E}}(y_a\otimes y_b\otimes y_c)$ to get $\widehat{\Pr}[h_{\text{new}}]$ and $\widehat{\mathbb{E}}(y_r|h_{\text{new}})$, $\forall r=\{a,b,c\}$.
 		\ENDIF
 		\ENDFOR
 		\ENDWHILE
 %\ENDFOR
 \end{algorithmic}
 \end{algorithm} %}

The divide-and-conquer local spectral parameter estimation is superior compared to popular EM-based method~\citep{choi2011learning}, which is slow and prone to local optima.
More importantly, EM can only be applied on a stable structure since it is a global update procedure.
Our proposed spectral learning method, in contrast, is applied locally over small groups of variables, and is a guaranteed learning with sufficient number of samples~\citep{anandkumar2012tensor}. Moreover, since we integrate structure and parameter learning, we avoid recomputing the same quantities, e.g. SVD computations are required both for structure estimation (for computing distances) and parameter estimation (for whitening the tensor). Combining these operations results in huge computational savings (see Section~\ref{sec:complexity} for the exact computational complexity of our method).

\begin{algorithm}[hbtp]
\caption{\small Merging and Alignment Correction (MAC)}\label{algo:pmac}
\begin{algorithmic}[1]
\REQUIRE \emph{Latent sub-trees} $\Adj_i$ for all internal nodes $i$.
%\REQUIRE Corresponding conditional means estimated in PLRG. 
%\ENSURE Structure and parameters of the entire latent tree graphical model.
\ENSURE Global latent tree $T$ structure and parameters.
\FOR {$\Adj_i$ and $\Adj_j$ in all the sub-trees}
\IF{there are common nodes between $\Adj_i$ and $\Adj_j$ }
\STATE Find the shortest path$(v_i,v_j;\Adj_i)$ between $v_i$ and $v_j$ on $\Adj_i$ and path$(v_i,v_j;\Adj_j)$ in $\Adj_j$;
\STATE Union the only conflicting path$(v_i,v_j;\Adj_i)$ \& path$(v_i,v_j;\Adj_j)$ according to equation~\eqref{eq:unionpath};
\STATE Attach other nodes in $\Adj_i$ and $\Adj_j$ to the union path; 
\STATE Perform alignment correction as described in Procedure~\ref{algo:alignment}.
\ENDIF
\ENDFOR
\end{algorithmic}
\end{algorithm}
\subsection{Merging and Alignment Correction}\label{sec:align}
We have so far learnt sub-trees  and  parameters over local groups of variables, where the groups are determined by the neighborhoods of the MST. The challenge now is to combine them to obtain a globally consistent estimate.
There are non-trivial obstacles to achieving this:
first, the constructed local sub-trees span  overlapping groups of observed nodes,  and possess conflicting paths. Second, local parameters need to be re-aligned as we merge the subtrees to obtain globally consistent estimates. To be precise,  different tensor decompositions lead to permutations of the hidden labels (columns permutations of the transition matrices) across triplets.  Thus, we need to find the permutation matrix best correcting the alignment of  hidden states of the transition matrices, so as to guarantee global consistency.

\textbf{Structure Union:} We now describe the procedure to merge the local structures. We merge them in pairs to obtain the final global latent tree. Recall that $\Adj_i$ denotes a sub-tree constructed locally over a group, whose leader is node $v_i$. Consider a pair of subtrees
 $\Adj_i$ and $\Adj_j$, whose group leaders $v_i$ and $v_j$ are neighbors on the MST. Since $v_i$ and $v_j$ are neighbors, both the sub-trees contain them, and have different paths between them (with hidden variables added).  Moreover, note that this is the only conflicting path in the two subtrees. We now describe how we can resolve this: in $\Adj_i$, let $h_1^i$ be the neighboring hidden node for $v_i$ and $h_2^i$ be the neighbor of $v_j$. There could be more hidden nodes between $h_1^i$ and $h_2^i$. Similarly, in $\Adj_i$, let $h_1^j$ and $h_2^j$ be the corresponding nodes in $\Adj_j$.  The shortest path between $v_i$ and $v_j$ in the two sub-trees are given as follows:
\begin{align}
\text{path}(v_i,v_j;\Adj_i) & := [v_i- h_1^i- \ldots- h^i_{2}-v_j]\\
\text{path}(v_i,v_j;\Adj_j) & := [v_i- h_1^j- \ldots- h^j_{2}-v_j]
\end{align}
Then the union path is formed as follows: 
\begin{align}
\label{eq:unionpath}
\text{merge}& (\text{path}(v_i,v_j;\Adj_i) ,\text{path}(v_i,v_j;\Adj_j)) \nonumber\\
& : = [v_i- h_1^i- \ldots- h^i_{2} - h^j_{1} \ldots h_2^j - v_j]
\end{align}In other words, we retain the immediate hidden neighbor of each group leader, and break the paths on the other end. For example in Figure~\ref{Fig:StructureLearning}(d1,d2), we have the path $v_3-h_1-v_5$ in $\Adj_3$ and path $v_3-h_3-h_2-v_5$ in $\Adj_5$. The resulting path is $v_3-h_1-h_3-h_2-v_5$, as see in Figure~\ref{Fig:StructureLearning}(e).
After the union of the conflicting paths, the other nodes are attached to the resultant latent tree. We present the pseudo code in Procedure~\ref{algo:pmac} in \href{https://drive.google.com/file/d/1ZkKVJjLrI1aMf-VevMJT1B1saXhtXcmK/view?usp=sharing}{Appendix}~\ref{appen:alignment}.
\begin{algorithm}[hbtp]
\caption{Parameter Alignment Correction \\ \textbf{(}$\mathbb{G}_r$ denotes reference group, $\mathbb{G}_o$ denotes the list of other groups, each group has a reference node denoted as $\mathcal{R}_l$, and the reference node in $\mathbb{G}_r$ is $\mathcal{R}_g$. The details on alignment at line 8 is in \href{https://drive.google.com/file/d/1ZkKVJjLrI1aMf-VevMJT1B1saXhtXcmK/view?usp=sharing}{Appendix}~\ref{appen:alignment}.\textbf{)}}\label{algo:alignment}
\begin{algorithmic}[1]
\REQUIRE Triplets and unaligned parameters estimated for these triplets, denoted as $\text{Trip}(y_i,y_j,y_k)$.
\ENSURE Aligned parameters for the entire latent tree $T$.
\STATE Select $\mathbb{G}_r$ which has \emph{sufficient children};
\STATE Select refer node $\mathcal{R}_g $ in $\mathbb{G}_r$;
\FORALL {a, b  in $\mathbb{G}_r$}
\STATE  Align $\text{Trip}_{\text{in}}(y_a,y_b, \mathcal{R}_g)$;
\ENDFOR
\FORALL {$i_g$ in $\mathbb{G}_o$}
	\STATE Select refer node  $\mathcal{R}_l $ in $\mathbb{G}_o$[$i_g$];
	\STATE Align $\text{Trip}_{\text{out}}(\mathcal{R}_g,  y_a, \mathcal{R}_l)$ and $\text{Trip}_{\text{out}}(\mathcal{R}_l, y_i, \mathcal{R}_g)$;
	\FORALL {i, j in $\mathbb{G}_o$[$i_g$]}
		\STATE Align $\text{Trip}(y_i,y_j,\mathcal{R}_l)$;
	\ENDFOR		
\ENDFOR
\end{algorithmic}
\end{algorithm}

\textbf{Parameter Alignment Correction:} As mentioned before, our parameter estimation is unsupervised, and therefore, columns of the estimated transition matrices may be permuted across different triplets over which tensor decomposition is carried out. The parameter estimation within the triplet is automatically acquired through the tensor decomposition technique, so that the alignment issue only arises across triplets. We refer to this as the alignment issue and it appears at various levels.

 There are two types of triplets, namely, \emph{in-group} and \emph{out-group} triplets. A triplet of nodes $\text{Trip}(y_i,y_j,y_l)$ is said to be \emph{in-group} (denoted by  $\text{Trip}_{\text{in}}(y_i,y_j,y_l)$ ) if
 its containing nodes share a joint node $h_k$ and there are no other hidden nodes in path($y_i$, $h_k$), path($y_j$, $h_k$) or path($y_l$, $h_k$).
 Otherwise, this triplet is \emph{out-group} denoted by $\text{Trip}_{\text{out}}(y_i,y_j,y_l)$.
 We define a group as \emph{sufficient children} group if it contains at least three \emph{in-group} nodes. %children for the joint node.
 
 %See Figure~\ref{Fig:StructureLearning}(d1) for an example of \emph{sufficient children} \emph{in-group estimation} with respect to $h_1$ and Figure~\ref{Fig:StructureLearning}(d2) \emph{insufficient children} \emph{in-group estimation} wrt $h_3$ and $h_2$.

%Within the thread, each local \emph{latent sub-tree} is divided into multiple \emph{in-groups}, so we need to align the parameters across these \emph{in-groups}. Additionally, within each \emph{in-group}, correction of alignment is needed as well. The alignment step is thus designed in various levels and is depicted in procedure~\ref{algo:alignment}.

Designing \emph{in-group} alignment correction with \emph{sufficient children} is relatively simple: we achieve this by including a local reference node for all the \emph{in-group} triplets. Thus, all the triplets are aligned with the reference node.  The alignment correction is more challenging if lacking \emph{sufficient children}. We propose \emph{out-group} alignment to solve this problem. We first assign one group as a \emph{reference group}, and the \emph{local reference node} in that \emph{reference group} becomes the \emph{global reference node}. In this way, we align all recovered transition matrices in the same order of hidden states as in the reference node. See Procedure~\ref{algo:pmac} and~\ref{algo:alignment} for merging the local structures and align the parameters from LRG local sub-trees.

\section{THEORETICAL GUARANTEES}\label{sec:complexity}
\textbf{Correctness of Proposed Parallel Algorithm: }We now provide the main result of this paper on global consistency for our method. %, despite the high degree of parallelism.

\begin{theorem}[Sample Complexity]\label{theorem:main}
	Given samples from an identifiable latent tree model, the proposed method consistently recovers the structure with $O(\log p)$ sample complexity and parameters with $O(\log k)$ sample complexity, where $p$ is the number of nodes and $k$ is the dimension of hidden variable which is usually small.
\end{theorem}
The proof sketch is in \href{https://drive.google.com/file/d/1ZkKVJjLrI1aMf-VevMJT1B1saXhtXcmK/view?usp=sharing}{Appendix}~\ref{appen:guarantee}. Lemma~\ref{lm:structure_robustness} and Lemma~\ref{lm:parameter_robustness} are used to prove Theorem~\ref{theorem:main}. Note that the sample complexity of our algorithm is dimension independent, therefore easily scalable to large $d$.

In order to understand the correctness of the structure learning part of our algorithm, the following Lemma~\ref{lm:structure_robustness} states a guaranteed consistency of the structure learning.
\begin{lemma}[Consistency of Structure Learning]\label{lm:structure_robustness}
	Let $\widehat{\mathcal{T}}^N$ be our estimated tree using $N$ number of examples, then for every $\eta>0$, if $N>C\log(p/\eta^{1/3})$ for some constant $C>0$, the error probability for structure reconstruction is upper bounded by $\eta$, i.e.,
	\begin{equation}
	\Pr(h(\widehat{\mathcal{T}}^N \neq \mathcal{T}))\le \eta
	\end{equation}
	where $h$ is a graph homomorphism -- a mapping between graphs that respects their structure.
\end{lemma}
{The multivariate information distance introduced in our paper enjoys the statistical efficiency in~\cite{anandkumar2011spectral} as  both methods require ``enough'' samples to get a confident estimation of the additive distance, which involves spectral decomposition of the empirical covariance. Although there seems to be a log difference, their algorithm requires ``multiplications'' of the metrics when testing quarterts, whereas we use ``summations'' of the metrics. Both analyses result in the same sample complexity. }

Lemma~\ref{lm:parameter_robustness} guarantees consistency of the parameter learning part of our algorithm.
{Tensor decomposition guarantees consistent estimation of the parameters with $O(\log k)$ examples, one of the key contributions compared to~\cite{choi2011learning} which uses EM.  }

% Statistical efficiency result of 
\textbf{Computational Complexity: }
Recall $d$ is the observable node dimension, $k$ is the hidden node dimension ($k \ll d$), $N$ is the number of samples, $p$ is the number of observable nodes, and $z$ is the number of non-zero elements in each sample. Let $\Gamma$ denote the maximum size of the groups, over which we operate the local recursive grouping procedure.  Thus, $\Gamma$ affects the degree of parallelism for our method. Recall that it is given by the neighborhoods on MST, i.e., $\Gamma : = \max_{i} \lvert\Nb[i;\text{MST}]\rvert$.   Below, we provide a bound on $\Gamma$.
\begin{lemma}\label{lem:MST}
The maximum size of neighborhoods on MST, denoted as $\Gamma$, satisfies
$
\Gamma \le \Delta^{1+\frac{u_{d}}{l_{d}}\delta},
$ where $\delta := \max_{i} \{\min_{j} \{\text{path}(v_i,v_j;\mathcal{T}) \} \} $ is the effective depth, $\Delta$ is the maximum degree of $\mathcal{T}$, and the $u_{d}$ and $l_{d}$ are the upper and lower bound of information distances between neighbors on $\mathcal{T}$.
\end{lemma}
Thus, we see that for many natural cases, where the degree and the depth in the latent tree are bounded (e.g. the hidden Markov model), and the parameters are mostly homogeneous (i.e., $u_d/l_d$ is small), the group sizes are bounded, leading to a high degree of parallelism. We summarize the computational complexity in Table~\ref{tab:computational_complexity}. Details can be found in \href{https://drive.google.com/file/d/1ZkKVJjLrI1aMf-VevMJT1B1saXhtXcmK/view?usp=sharing}{Appendix}~\ref{appen:compuComplex}. %Overall we see that our algorithm is bulk-asynchronous parallel.

 {The reason we emphasize parallel complexity is to show the high ``degree of parallelism'' of our algorithm, one of the main advantages of our algorithm over a sequential implementation. Given two algorithms with the same serial complexity, we prefer the one with higher degree of parallelism as it could be made more efficient.  }
\begin{table}[htbp]
   \centering
   \small{
\begin{tabular}{@{} l|l|l@{}}
Algorithm Step & Time per worker & {Degree of parallelism}\\
\hline
Distance Est. &  $O( N z +  d + k^3 )$ & $O(p^2)$\\ %Information Distance Estimation
MST & $O(\log p)$ & $O(p^2)$\\ %Structure: Minimum Spanning Tree
LRG & $O(\Gamma^3)$ & $O(p/ \Gamma)$\\ %Structure: Local Recursive Grouping
Tensor Decomp. & $O(\Gamma k^3 + \Gamma d k^2)$ & $O(p/ \Gamma)$\\ %Parameter: Tensor Decomposition
Merging step& $O(d k^2)$ & $O(p/ \Gamma)$\\ %Merging and Alignment Correction
\end{tabular}
}
\caption{\small{Worst-case computational complexity of our algorithm. The total complexity is the product of the time per work and degree of parallelism.}}\label{tab:computational_complexity}%The times refer to the worst-case complexity. According to Lemma~\ref{lem:MST}, $\Gamma$ is small if $\Delta$ and $\delta$ are small. }
\end{table}

\section{EXPERIMENTS}% - Healthcare Data Analysis}
\label{sec:implementation}
%See Appendix~\ref{appen:synthetic} for synthetic experiments and some real experiments of NIPS dataset from the UCI bag-of-words repository.
%\fhcomment{Our parallelizable algorithm is faster and suffers from \emph{no performance loss} compared with the sequential approach, due to the global consistency guarantee in our theorem. Therefore, we didn't include this comparison in our experiment section, although verified with experiments. }

%\fhcomment{ \textbf{Comparison with other statistical models:} To the best of our knowledge, the existing statistical models with structure and parameter estimation guarantees are the sequential counterpart of our parallel algorithm. Our approach incurs no performance loss by speeding up the process. Therefore, our algorithm achieves the state-of-the-art accuracy and parallel computational complexity. }

\begin{figure}
	\centering
	\psfrag{samples}[c]{{samples}}
	\psfrag{err}[c]{{error}}
	\psfrag{Structure Error long}{{Structure Error}}
	\psfrag{EM}{{CLRG-EM}}
	\psfrag{TF}{{Proposed}}
	\includegraphics[height =1in]{\fighome/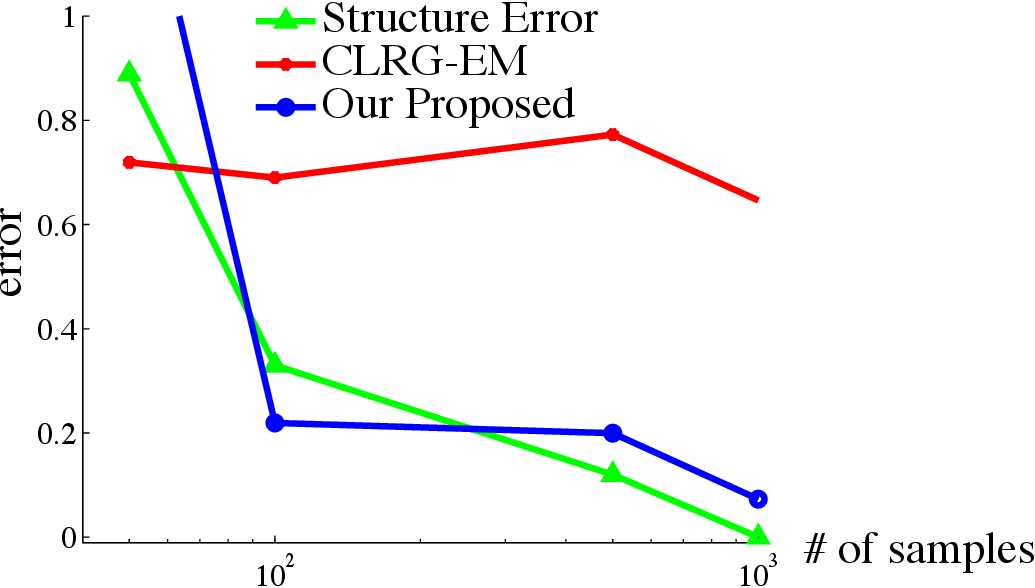}
	\caption{Comparison of structure and parameters recovery error between CLRG structure learning with EM parameter estimation and our algorithm. Structure errors (green) are the same for the two approaches. Proposed (blue) parameter estimation performs better than EM (red) ($d=2$, $p=9$). }\label{fig:Para}
\end{figure}

%\fhcomment{Evaluating the performance of tree structure recovery is indeed nontrivial as the \emph{number} and \emph{location} of the hidden variables vary as well as the depths. Two trees may be similar but may look different due to the difference that how refined the hierarchical clusterings are.  RF is a well defined metric for evaluating the difference of two tree structures. Similar to other distance metrics, smaller RF indicates the recovered structure being closer to the ground-truth. 0 is optimal. }
\textbf{Synthetic experiments.} We compare our method with CLRG-EM~\citep{choi2011learning}, where a sequential learning procedure is carried out for binary variables $(d=2)$ and parameter learning is carried out via EM. We consider a latent tree model over nine observed variables and four hidden nodes. We restart EM 20 times, and select the best result. The results are in Figure~\ref{fig:Para}. 
We measure the structure recovery error via $\left\{\sum\limits_{i=1}^{\lvert \widehat{G}\rvert}\min_j \lvert\widehat{G}_i\not\in G_j\rvert/{\lvert \widehat{G}_i\rvert}\right \}/{\lvert \widehat{G}\rvert}$, where $G$ and $\widehat{G}$ are the ground-truth and recovered categories.
We measure the parameter recovery error via $\mathcal{E} = \| A - \hat{A} \Pi \|_F$, where $A$ is the true parameter and $\hat{A}$ is the estimated parameter and $\Pi$ is a suitable permutation matrix that aligns the columns of $\hat{A}$ with $A$ so that they have minimum distance. $\Pi$ is greedily calculated. It is shown that as number of samples increases, both methods recover the structure correctly, as predicted by the theory. However, EM is stuck in local optima and fails to recover the true parameters, while the   tensor decomposition correctly recovers the true parameters.
We present the high performance of our algorithm on the large $p$ and $d$ regime where CLRG-EM is slow and easily stuck in local optima, in Table~\ref{tab:synthetic}. We achieve efficient running times with good accuracy for structure and parameter estimation.

\begin{table}[h]
	\centering
	\begin{tabular}{@{} l|l|l|l|l|l @{}}
		    &     &     & Struct & Param & Running \\
		$d$ & $p$ & $N$ & Error & Error & Time(s)\\
		\hline
		10 		&  	9 & 50K & 0 	& 0.0104 	& 3.8	\\ %0.8
		100 	& 	9 & 50K & 0		& 0.0967 	& 4.4	\\ %0.8
		1000 	&  	9 & 50K & 0     & 0.1014 	& 5.1	\\%0.8
		10,000	&	9 & 50K & 0		& 0.0917	& 29.9	\\ %0.98
		100,000 &   9 & 50k & 0		& 0.0812	& 56.5	\\%0.998
		\hline
		100 & 9 	& 50K & 0  		& 0.0967 	& 10.9	\\
		100 & 81 	& 50K & 0.06 	& 0.1814 	& 323.7	\\
		100 & 729 	& 50K & 0.16	& 0.1913	& 4220.1\\
	\end{tabular}
	\caption{Algorithm performance in large $d$ and $p$ regime where CLRG-EM is not amendable and in large $p$ regime where CLRG-EM is slow and easily stuck in local optima.}   \label{tab:synthetic}
\end{table}
\textbf{Real data experiments on NIPS and NY Times.} Real datasets experiments which estimate a hierarchical structure over words are implemented and presented in Appendix~\ref{appen:realdata} Figure~\ref{fig:global-nips}, ~\ref{fig:nips_plots} and~\ref{fig:nytimes}. The relationships among the words discovered by our algorithm match intuition. For example in Figure~\ref{fig:nytimes}, govern and secur are grouped together whereas movi, studio and produc are clustered.
%We keep other parameters fixed while varying two of them. 
%We perform experiments on a synthetic tree with observable dimension $d=1,000,000$, number of observable nodes $p=3$, hidden dimension $k=3$ and number of samples $N=1000$ which is similar to community detection setting of \cite{DBLP:journals/corr/HuangNHVA13}. We note that the recovery of the structure and the parameters is done in $10.6$ seconds.

\textbf{Healthcare data analysis.} We demonstrate that our algorithm works for challenging tasks such as healthcare analytics in Appendix~\ref{app:healthcare}.
Our goal is to discover a disease hierarchy based on their co-occurring relationships in the patient records. In general, longitudinal patient records store the diagnosed diseases on patients over time, where the diseases are encoded with International Classification of Diseases (ICD) code. 
We use two large patient datasets (\textbf{MIMIC2} and \textbf{CMS}) of different sizes with respect to the number of samples, variables and dimensionality. The details of the datasets are discussed in \href{https://drive.google.com/file/d/1ZkKVJjLrI1aMf-VevMJT1B1saXhtXcmK/view?usp=sharing}{Appendix}~\ref{sec:data_describ}.  The goal is to learn the latent nodes and the disease hierarchy and associated parameters from data. We validate the resulting disease hierarchy both quantitatively and qualitatively, and verify the scalability of our algorithm in Appendix~\ref{app:healthcare}.

\section*{Acknowledgement}
\vspace{-1em}
Huang is supported by startup fund from Department of Computer Science, University of Maryland, National Science Foundation IIS-1850220 CRII Award 030742-00001,  and  Adobe, Capital One and JP Morgan faculty fellowships.
Sun is supported by the National Science Foundation award IIS-1418511, CCF-1533768 and IIS-1838042, the National Institute of Health award 1R01MD011682-01 and R56HL138415.
Anandkumar is supported in part by Bren endowed chair, Darpa PAI, Raytheon, and Microsoft, Google and Adobe faculty fellowships.
\newpage
{%\bibliographystyle{alpha}
\bibliographystyle{abbrvnat}
\setcitestyle{authoryear,open={((},close={))}}
\bibliography{\bibhome/latentTree_ref}
}
\newpage
\appendix
\onecolumn
\centerline{\Large \textbf{Appendix:\mytitle}}
%!TEX root = 0_uai_latent_tree_main.tex

\section{RELATED WORKS}
There has been widespread interest in developing distributed learning techniques, e.g.~the recent works of~\cite{smola2010architecture} and~\cite{2013arXiv1312.7869W}.
These works consider parameter estimation via likelihood-based optimizations such as Gibbs sampling, while our method involves more challenging tasks where both the structure and the parameters are estimated.
%Various works have established that stochastic gradient descent can be parallelized efficiently, e.g.~\cite{niu2011hogwild} establish a lock-free guarantee for hog-wild stochastic gradient descent.
%On the other hand, most previous structure learning procedures for graphical models cannot be easily parallelized.
Simple methods such as local neighborhood selection through $\ell_1$-regularization~\citep{Mei06} or local conditional independence testing~\citep{AnandkumarTanWillsky:Ising11} can be parallelized, but  these methods do not  incorporate hidden variables.
Finally, note that the latent tree models provide a statistical description, in addition to revealing the hierarchy.
In contrast, hierarchical clustering techniques are not based on a statistical model~\citep{krishnamurthy2012efficient} and cannot provide valuable information such as the level of correlation between observed and hidden variables.

Both~\cite{chang1991reconstruction} and~\cite{chang1996full} provide detailed identifiability analyses of latent trees and the former~\citep{chang1991reconstruction} introduces maximum likelihood estimation method, different from our guaranteed parallel spectral method using tensor decomposition.

\cite{p2011spectral,song2011kernel,song2014nonparametric} are closely related to our paper. The~\cite{p2011spectral} paper does not address the structure learning. The sample complexity is polynomial in $k$, whereas ours is logarithmic in $k$. 
The~\cite{song2014nonparametric} paper provides guarantees for structure learning, but not for parameter estimation. 
The~\cite{song2011kernel} paper does not provide sample complexity theorem or analysis for recovering the latent tree structure or parameter with provable guarantees.

%\fhcomment{The main novel contributions of this paper (which are missing in~\cite{choi2011learning}) are (1) to propose a guaranteed hierarchical parameter estimation mechanism for latent tree models and (2) to provide a local and parallel algorithm with global consistency guarantee. }
%\onecolumn

\section{SYNTHETIC EXPERIMENTS} \label{appen:synthetic}

\subsection{Machine Setup}
\textbf{Setup }Experiments are conducted on a server running the Red Hat Enterprise 6.6 with 64 AMD Opteron processors and 265 GBRAM. The program is written in C++, coupled with the multi-threading capabilities of the OpenMP environment~\citep{OMP} (version 1.8.1).
We use the Eigen toolkit\footnote{\scriptsize{\url{http://eigen.tuxfamily.org/index.php?title=Main_Page}}} where BLAS operations are incorporated. For SVDs
of large  matrices, we use randomized projection methods~\citep{gittens2013revisiting} as described in Appendix~\ref{apdx:svd}.

\subsection{Synthetic Results:} We compare our method with the implementation of~\cite{choi2011learning}, where a serial learning procedure is carried out for binary variables $(d=2)$ and parameter learning is carried out via EM. We consider a latent tree model over nine observed variables four hidden nodes. We restart EM 20 times, and select the best result. The results are in Figure~\ref{fig:Para}. 

We measure the structure recovery error via $\left\{\sum\limits_{i=1}^{\lvert \widehat{G}\rvert}\min_j \lvert\widehat{G}_i\not\in G_j\rvert/{\lvert \widehat{G}_i\rvert}\right \}/{\lvert \widehat{G}\rvert}$, where $G$ and $\widehat{G}$ are the ground-truth and recovered categories.
We measure the parameter recovery error via $\mathcal{E} = \| A - \hat{A} \Pi \|_F$, where $A$ is the true parameter and $\hat{A}$ is the estimated parameter and $\Pi$ is a suitable permutation matrix that aligns the columns of $\hat{A}$ with $A$ so that they have minimum distance. $\Pi$ is greedily calculated. It is shown that as number of samples increases, both methods recover the structure correctly, as predicted by the theory. However, EM is stuck in local optima and fails to recover the true parameters, while the   tensor decomposition correctly recovers the true parameters.

%\begin{figure}[H]
%	\centering
%	\psfrag{samples}[c]{{samples}}
%	\psfrag{err}[c]{{error}}
%	\psfrag{Structure Error long}{{Structure Error}}
%	\psfrag{EM}{{CLRG-EM}}
%	\psfrag{TF}{{Proposed}}
%	\includegraphics[width=0.4\textwidth]{\fighome/ParaStruct.eps}
%	\caption{Comparison of structure and parameters recovery error between CLRG structure learning with EM parameter estimation and our algorithm. Structure errors (green) are the same for the two approaches. Proposed (blue) parameter estimation performs better than EM (red) ($d=2$, $p=9$). }\label{fig:Para}
%\end{figure}

We then present our experimental results focusing on the large $p$ and even larger $d$ regime. See Table~\ref{tab:synthetic}. We achieve efficient running times with good accuracy for structure and parameter estimation.

%\begin{table}[h]
%	\centering
%	\begin{tabular}{@{} l|l|l|l|l|l @{}}
%		$d$ & $p$ & $N$ & Struct Error & Param Error & Running Time(s)\\
%		\hline
%		10 		&  	9 & 50K & 0 	& 0.0104 	& 3.8	\\ %0.8
%		100 	& 	9 & 50K & 0		& 0.0967 	& 4.4	\\ %0.8
%		1000 	&  	9 & 50K & 0     & 0.1014 	& 5.1	\\%0.8
%		10,000	&	9 & 50K & 0		& 0.0917	& 29.9	\\ %0.98
%		100,000 &   9 & 50k & 0		& 0.0812	& 56.5	\\%0.998
%		\hline
%		100 & 9 	& 50K & 0  		& 0.0967 	& 10.9	\\
%		100 & 81 	& 50K & 0.06 	& 0.1814 	& 323.7	\\
%		100 & 729 	& 50K & 0.16	& 0.1913	& 4220.1\\
%	\end{tabular}
%	\caption{Algorithm performance in large $d$ and $p$ regime where CLRG-EM is not amendable and in large $p$ regime where CLRG-EM is slow and easily stuck in local optima.}   \label{tab:synthetic}
%\end{table}

% We keep other parameters fixed while varying two of them. 
We perform experiments on a synthetic tree with observable dimension $d=1,000,000$, number of observable nodes $p=3$, hidden dimension $k=3$ and number of samples $N=1000$ which is similar to community detection setting of \cite{DBLP:journals/corr/HuangNHVA13}. We note that the recovery of the structure and the parameters is done in $10.6$ seconds.

\section{
EXPERIMENTS ON NIPS AND NY TIMES DATASET}\label{appen:realdata}
\begin{figure*}[htbp]
			{\begin{minipage}{0.5\textwidth}\centering\label{Fig:algorithm}
					\small \centering
					\includegraphics[width=\textwidth]{\fighome/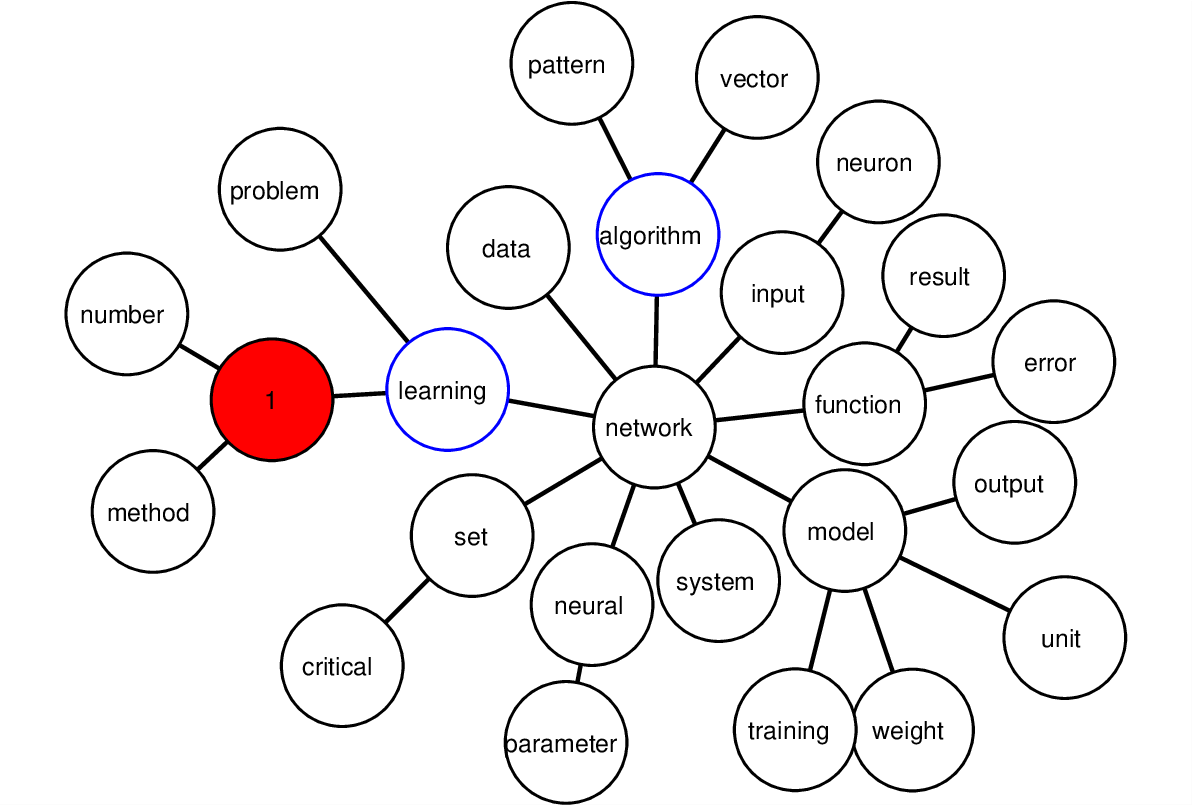}
					
					%(a)	Algorithm \& Learning
				\end{minipage}}
				\hfil
				{\begin{minipage}{0.5\textwidth}\centering\label{Fig:eigenvalues}
						\small \centering
						\psfrag{data}[c]{DATA}
						\includegraphics[width=\textwidth]{\fighome/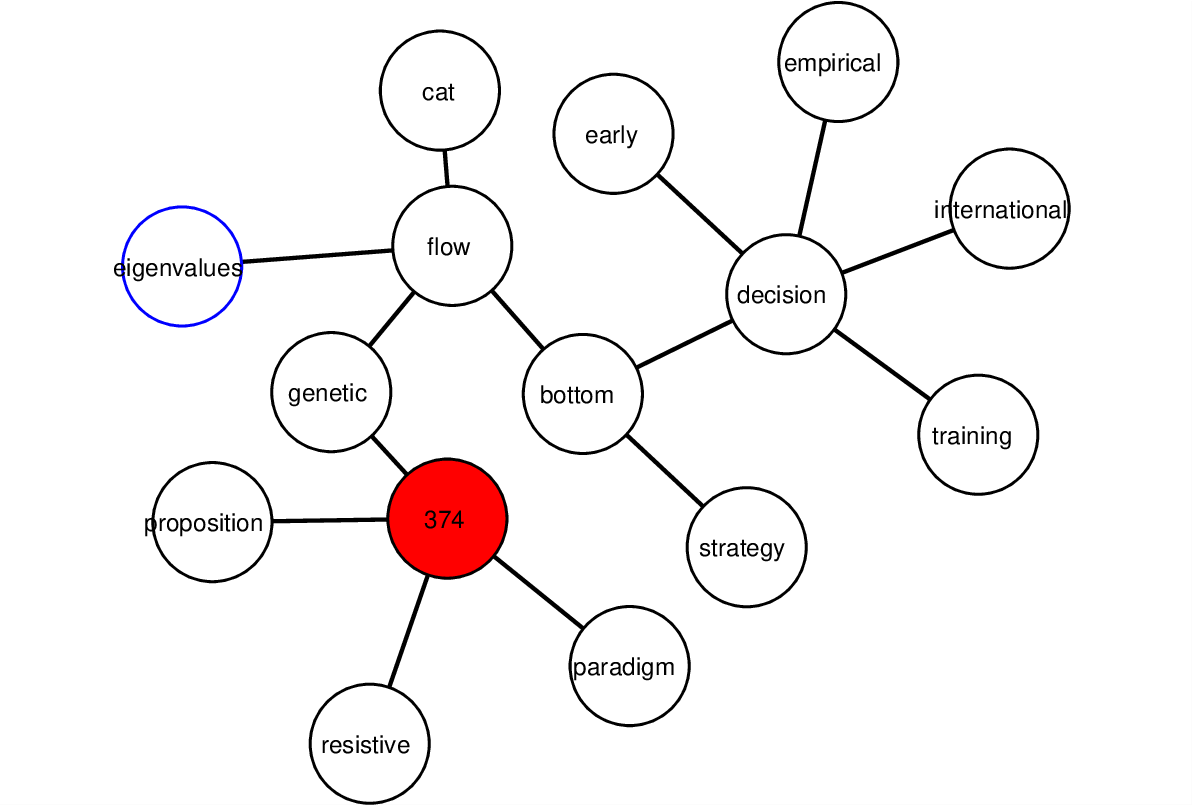}
						
						%(b)	Eigenvalues
					\end{minipage}}
					\hfil
					{\begin{minipage}{0.5\textwidth}\centering\label{Fig:exponentially}
							\small \centering
							\includegraphics[width=\textwidth]{\fighome/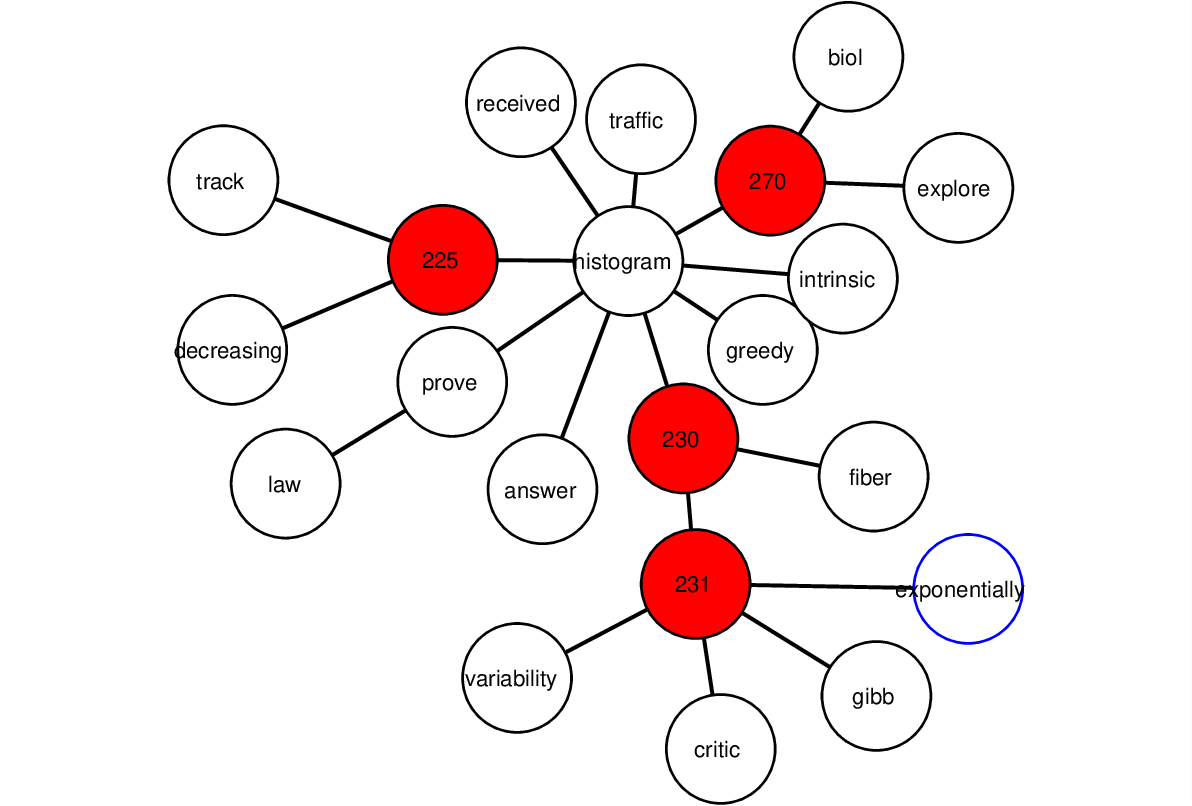}%t_vs_samples_2.eps
							
							%(c)	Exponentially
						\end{minipage}}
						\hfil
						\hspace{-1em}
						{\begin{minipage}{0.5\textwidth}\centering\label{Fig:latent}
								\small \centering
								\includegraphics[width=\textwidth]{\fighome/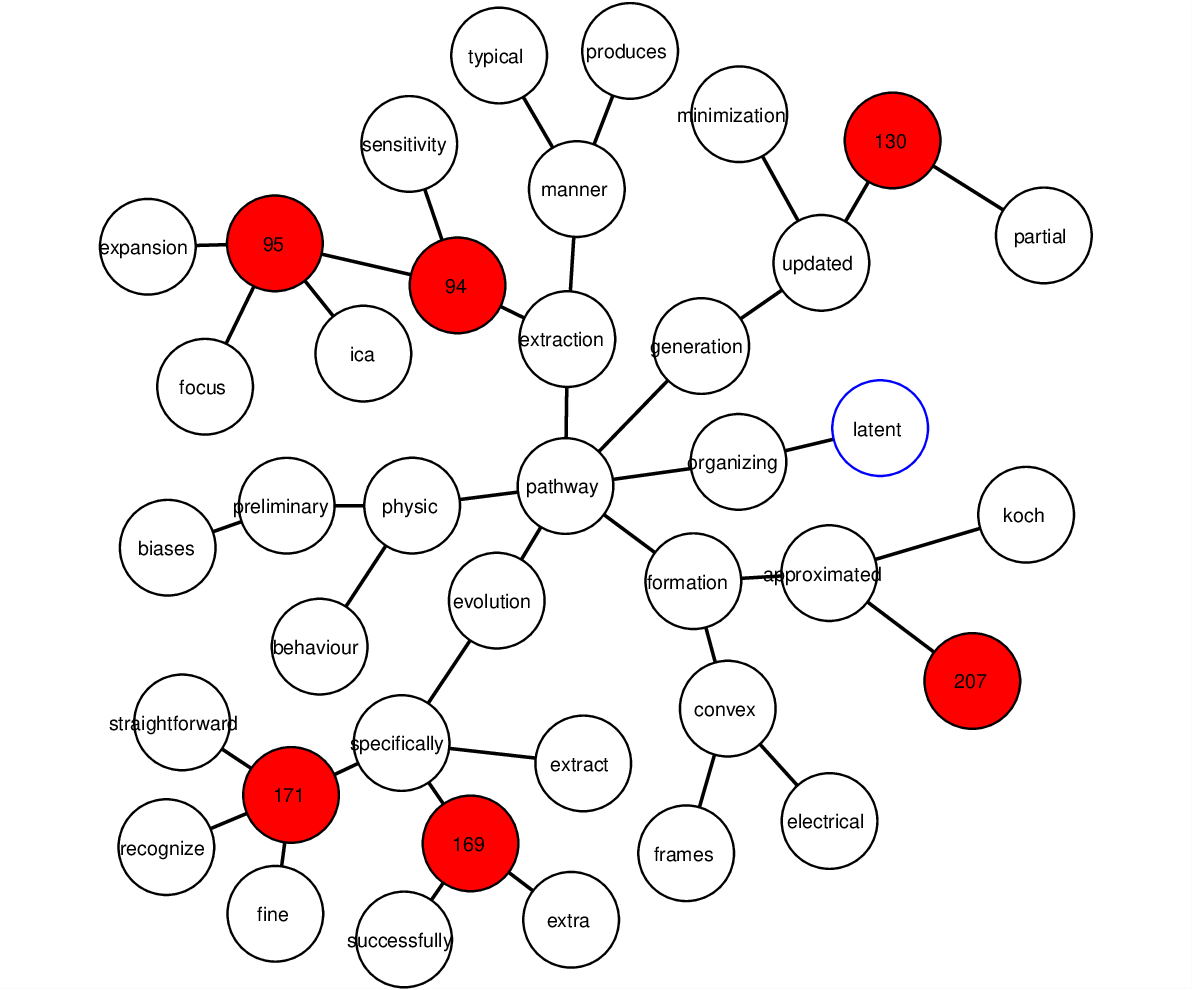}%error_vs_samples_slog.eps
								
								%(d)	Latent
							\end{minipage}}
							\caption{Extended neighborhoods of some of the words estimated from the NIPS dataset. We run our algorithm for the top 5000 keywords global and local hierarchical structures.}
							\label{fig:nips_plots}
						\end{figure*}

We use the entire NIPS dataset from the UCI bag-of-words repository.  This consists of $1500$ documents and $12419$ words in the vocabulary. We estimate the hierarchical structure of the whole corpus in only 15332.4 seconds (4 hours). Additionally, we focus on the top $5000$ most frequently appeared keywords and illustrate the local structures in Figure~\ref{fig:nips_plots}.  The running time for the subset is only 1164.4 seconds (20 minutes). 

\begin{figure}[htbp]
							\centering
							\includegraphics[width=0.6\textwidth,height=2in]{\fighome/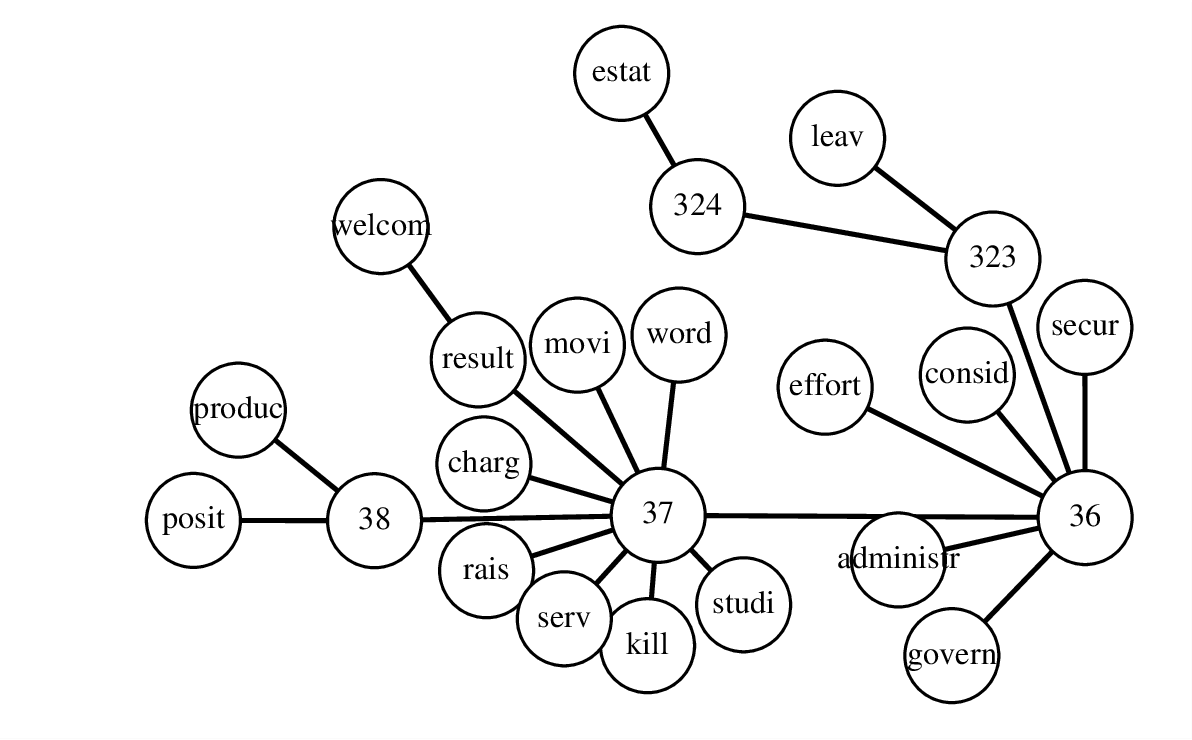} % requires the graphicx package
							\caption{A local view of recovered NY Times keywords structure}
							\label{fig:nytimes}
						\end{figure}

We also use the NY Times dataset from the UCI bag-of-words repository. % after carrying out the standard stemming procedure.
This consists of $300000$ documents and $102660$ words in the vocabulary. Our algorithm estimates the hierarchical structure of $3000$ most frequently appeared keywords in the NY Times dataset in only 107.8 seconds. Note that $d = 2$, since we consider the occurrence of a word in a document as a binary variable. Below is a subset of the keywords graph we estimated in Figure~\ref{fig:nytimes}. 
We note that the relationships the among the words in Figure~\ref{fig:nytimes} match intuition. For example, govern and secur are grouped together whereas movi, studio and produc are grouped together.  The numbers represent hidden nodes.

\begin{figure*}[htbp]
	{\begin{minipage}{0.7\textwidth}
			\centering
			\includegraphics{\fighome/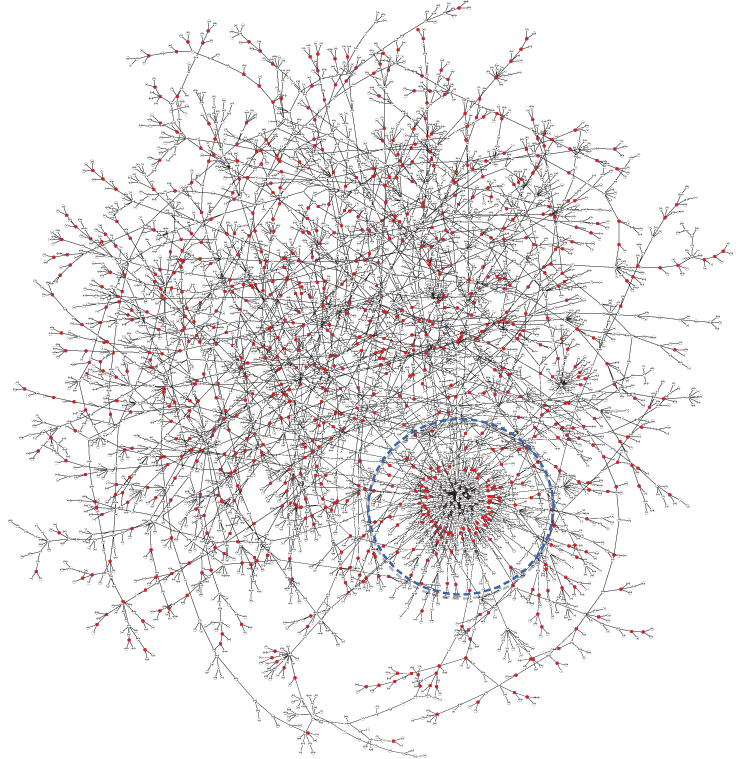} % requires the graphicx package
			
			Recovered Latent Tree
		\end{minipage}}
		\hfil
		{
			\begin{minipage}{0.2\textwidth}
				\centering
				\label{Fig:training}
				\small \centering
				\includegraphics[width=\textwidth]{\fighome/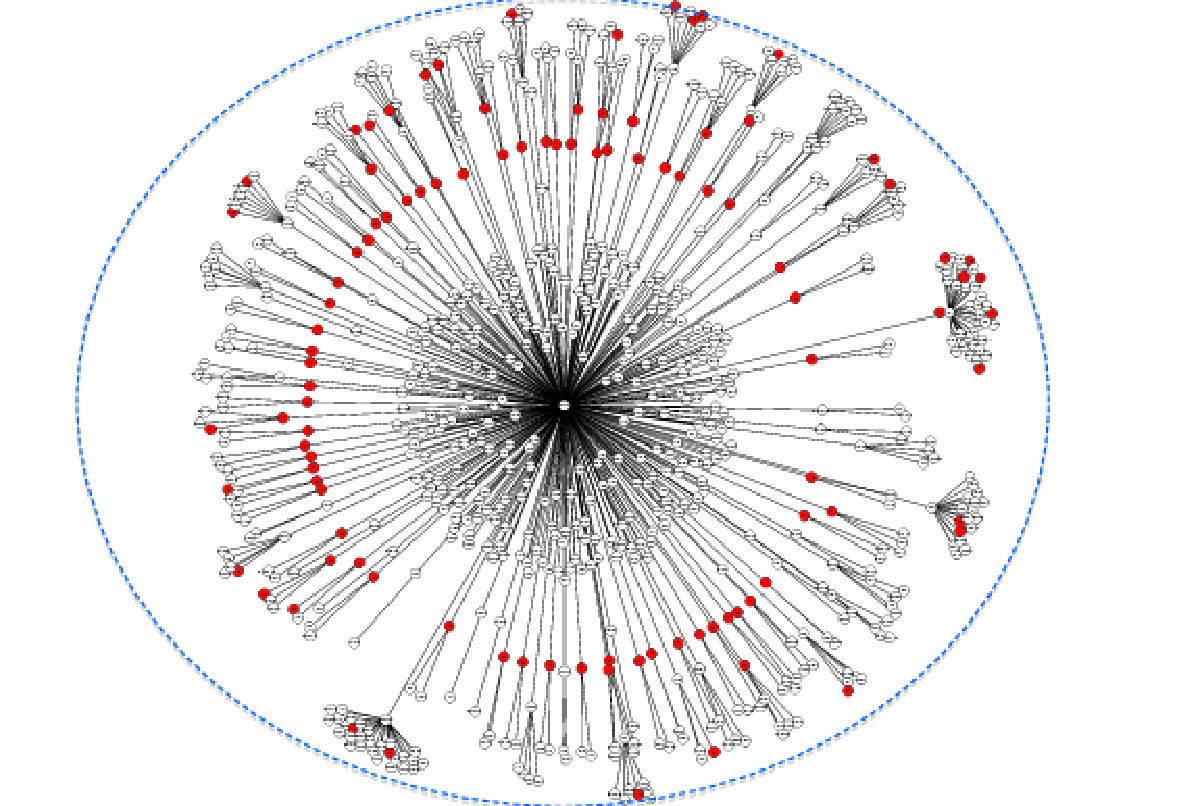}
				
				``Training'' Neighborhood as in the blue dashed circle
			\end{minipage}}
			\caption{Estimated NIPS top 5000 keywords global hierarchical structures. Red nodes are latent nodes introduced. The blue dash circle in the global hierarchical structure is zoomed in which is the neighborhood of word ``training''.}\label{fig:global-nips}
		\end{figure*}
		
						%\section{Hidden Node Estimation: Least Squares Approximation}\label{appen:leastsquare}
						%We approximate the posterior over a hidden node using a least squares approximation. Consider a hidden node $h$ with children $x_i, \forall i \in [l]$ with transition matrix from $h$ to $x_i$ denoted by $\zeta_i$.
						% We have
						%$\mathbb{E}[x_i] = \zeta_i \mathbb{E}[h] + \epsilon_i.
						%$
						%%\[
						%%\mathbb{E}[x_i] = E_i \mathbb{E}[h] + \epsilon_i \text{ with }
						%%\epsilon_i \sim \mathcal{N}(0, 1)
						%%\]
						%%A least square approach to this mean estimation of hidden parent is thus as follows. 
						% Now, we concatenate the samples $x_i$ to obtain $C\in \mathbb{R}^{l\times d}$ and concatenate the transition matrices $\zeta_i$ to obtain a $\zeta \in \mathbb{R}^{ld \times k}$. Hence, we estimate the mean of the hidden node by
						%$
						%\mathbb{E}[h] = \zeta^\dag C.
						%$

\section{EXPERIMENTS ON HEALTHCARE DATASETS}\label{app:healthcare}
%Qualitatively, we display our recovered tree structures for both the datasets with varying $d$. 
\textbf{Quantitative Analysis } We evaluate our resulting hierarchy with a ground truth tree,  based on medical knowledge\footnote{The ground truth tree is the PheWAS hierarchy provided in the clinical study~\citep{Denny:pheWAS}}. We compare our results with a baseline: the agglomerative clustering.   Evaluating the performance of tree structure recovery is indeed nontrivial as the \emph{number} and \emph{location} of the hidden variables vary as well as the depths. Two trees may be similar but may look different due to the difference that how refined the hierarchical clusterings are. The standard Robinson Foulds (RF) metric~\citep{robinson1981comparison}(between our estimated latent tree and the ground truth tree) is computed to evaluate the structure recovery in Table~\ref{tab:RFmetric}. RF is a well defined metric for evaluating the difference of two tree structures. Similar to other distance metrics, smaller RF indicates the recovered structure being closer to the ground-truth.The smaller the metric is, the better the recovered tree is.  
The proposed method is slightly better than the baseline and the advantage increases with more nodes. 
However, our proposed method provides an efficient probabilistic graphical model that can support general inference which is not feasible with the baseline.

%The recovered tree structure was compared to the ground truth tree, representing memberships of diseases into disease groups as used in the PheWAS encoding, which groups similar diseases by similarity~\citep{Denny:pheWAS}.  
\begin{table}[h]
\centering
\begin{tabular}{  c | c | c |c}
Data 		& $p$ 	&    RF(agglo.) 		& RF(proposed) \\
\hline
MIMIC2 	& 	163 & 0.0061	&  0.0061\\
CMS   	&	168 & 0.0060	&  0.0059\\
MIMIC2 	& 	952 & 0.0060	&  0.0011 \\
\end{tabular}
\caption{Robinson Foulds (RF) metric compared with the ``ground-truth'' tree for both MIMIC2 and CMS dataset. Our proposed results are better as we increase the number of nodes. } \label{tab:RFmetric}
\end{table}

\textbf{Qualitative analysis } 
Here we report the results from the 2-dimensional case (i.e., observed variable is binary). 
In Figure \ref{Fig:tree_mimic2_1} in appendix~\ref{sec:experimental_result}, we show a portion of the learned tree using the MIMIC2 healthcare data. The yellow nodes are latent nodes from the learned subtrees while the blue nodes represent observed nodes(diagnosis codes) in the original dataset.  Diagnoses that are similar were generally grouped together. For example, many neoplastic diseases were grouped under the same latent node (node 1135). While some dissimilar diseases were grouped together, there usually exists a known or plausible association of the diseases in the clinical setting. For example, in Figure \ref{Fig:tree_mimic2_1} in appendix~\ref{sec:experimental_result}, clotting-related diseases and altered mental status were grouped under the same latent node as several neoplasms. This may reflect the fact that altered mental status and clotting conditions such as thrombophlebitis can occur as complications of neoplastic diseases~\citep{Falanga:clotCancer}. The association of malignant neoplasms of prostate and colon polyps, two common cancers in males, is captured under latent node 1136~\citep{us2014united}.

For both the MIMIC2 and CMS datasets, we performed a qualitative comparison of the resulting trees while varying the hidden dimension $k$ for the algorithm. The resulting trees for different values of $k$ did not exhibit significant differences. This implies that our algorithm is robust with different choices of hidden dimensions. The estimated model parameters are also robust  for different values of $k$ based on the results.

\textbf{Scalability } Our algorithm is scalable w.r.t.~varying characteristics
of the input data. First, it can handle
a large number of patients efficiently,
as shown in Figure~\ref{Fig: scale}(a). % \jpcomment{To be completed}
%Our algorithm is scalable w.r.t.~varying characteristics of the input data. 
It has a linear scaling behavior as we vary the number observed
nodes, as shown in Figure~\ref{Fig: scale}(b).
Furthermore, even in cases where the number of observed
variables is large, our method maintains an almost linear scale-up as we
vary the computational power available, as shown in Figure~\ref{Fig: scale}(c).
So, by providing the respective resources, our algorithm is practical
under any variation of the input data characteristics.

\subsection{Data Description}\label{sec:data_describ}
\emph{(1) MIMIC2:} The MIMIC2 dataset record disease history of 29,862 patients where a overall of 314,647 diagnostic events over time representing 5675 diseases are logged. We consider patients as samples and groups of diseases as variables. We analyze and compare the results by varying the group size (therefore varying $d$ and $p$).

\emph{(2) CMS:} The CMS dataset includes 1.6 million patients, for whom 15.8 million medical encounter events are logged. Across all events, 11,434 distinct diseases (represented by ICD codes) are logged.  We consider patients as samples and groups of diseases as variables. We consider specific diseases within each group as dimensions. We analyze and compare the results by varying the group size (therefore varying $d$ and $p$). 
While the MIMIC2 dataset and CMS dataset both contain logged diagnostic events, the larger volume of data in CMS provides an opportunity for testing the algorithm's scalability. We qualitatively evaluate biological implications on MIMIC2 and quantitatively evaluate algorithm performance and scalability on CMS.

To learn the disease hierarchy from data,  we also leverage some existing domain knowledge about diseases. In particular, we use an existing mapping between 
ICD codes and higher-level  Phenome-wide Association Study (PheWAS) codes~\citep{Denny:pheWAS}. We use (about 200) PheWAS codes as observed nodes and the observed node dimension is set to be binary ($d=2$) or the maximum number of ICD codes within a pheWAS code ($d=31$). 
\subsection{Results}\label{sec:experimental_result}

{\textbf{Case d =31:} }
We learn a tree from the MIMIC2 dataset, in which we grouped diseases into 163 pheWAS codes and up to 31 dimensions per variable. % \rccomment{should we put the value of K used?} \fhcomment{That's fine as we say the results are robust with respect to k}.
Figure \ref{Fig:tree_mimic2_2} in appendix~\ref{sec:experimental_result} shows a portion of the learned tree of four subtrees which all reflect similar diseases relating to trauma. A majority of the learned subtrees reflected clinically meaningful concepts, in that related and commonly co-occurring diseases tended to group together in the same subtrees or in nearby subtrees. 

We also learn the disease tree from the larger CMS dataset, in which we group diseases into 168 variables and up to 31 dimensions per variable. Similar to the case from the MIMIC2 dataset, a majority of learned subtrees reflected clinically meaningful concepts. 

\begin{figure}[htbp]
	\includegraphics[width=\columnwidth]{\fighome/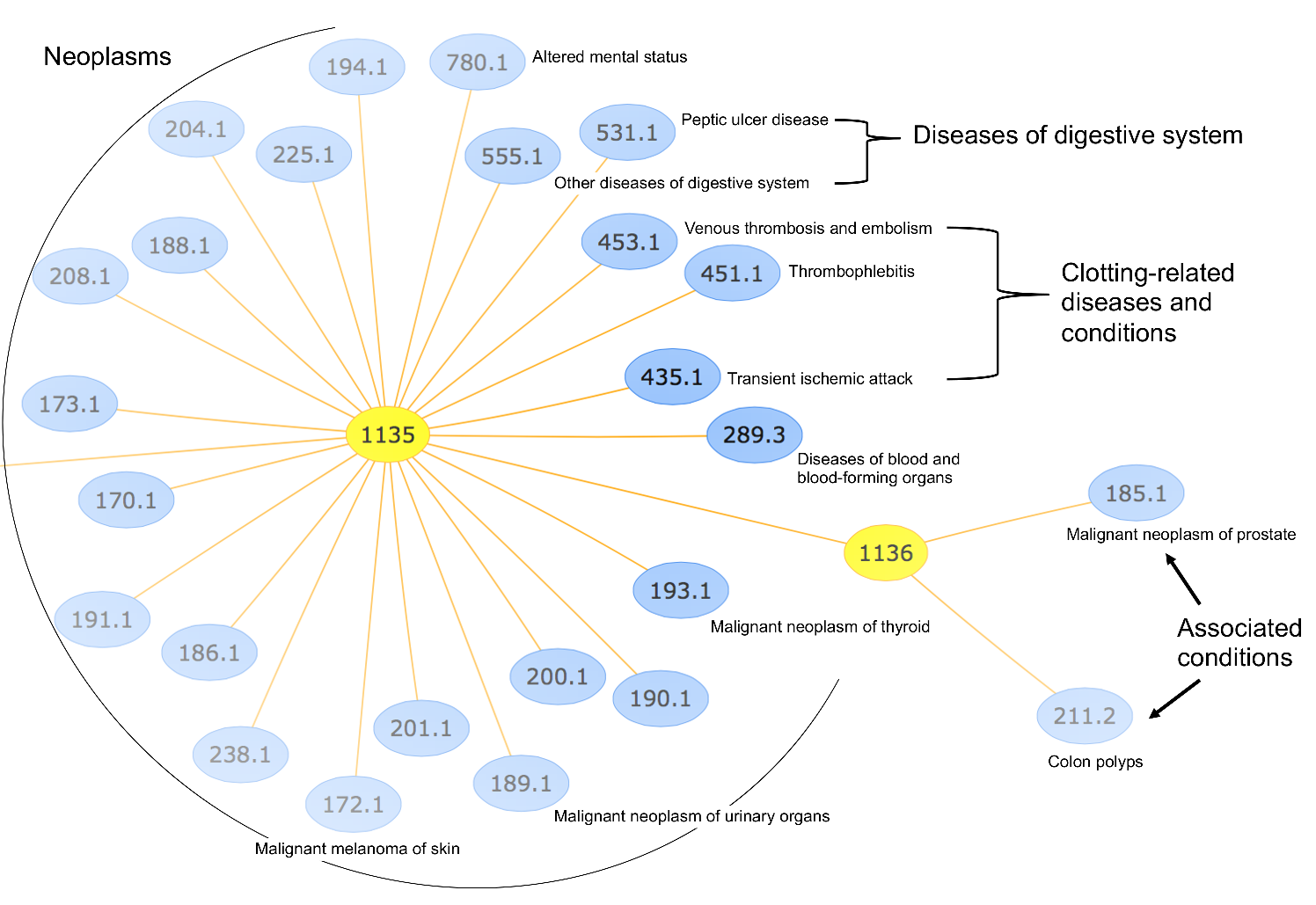}
	\vspace{-1em}
	\caption{An example of two subtrees which represent groups of similar diseases which may commonly co-occur. Nodes colored yellow are latent nodes from learned subtrees.} \label{Fig:tree_mimic2_1}
\end{figure}

\begin{figure}[htb]
\begin{center}
\includegraphics[width=\columnwidth]{\fighome/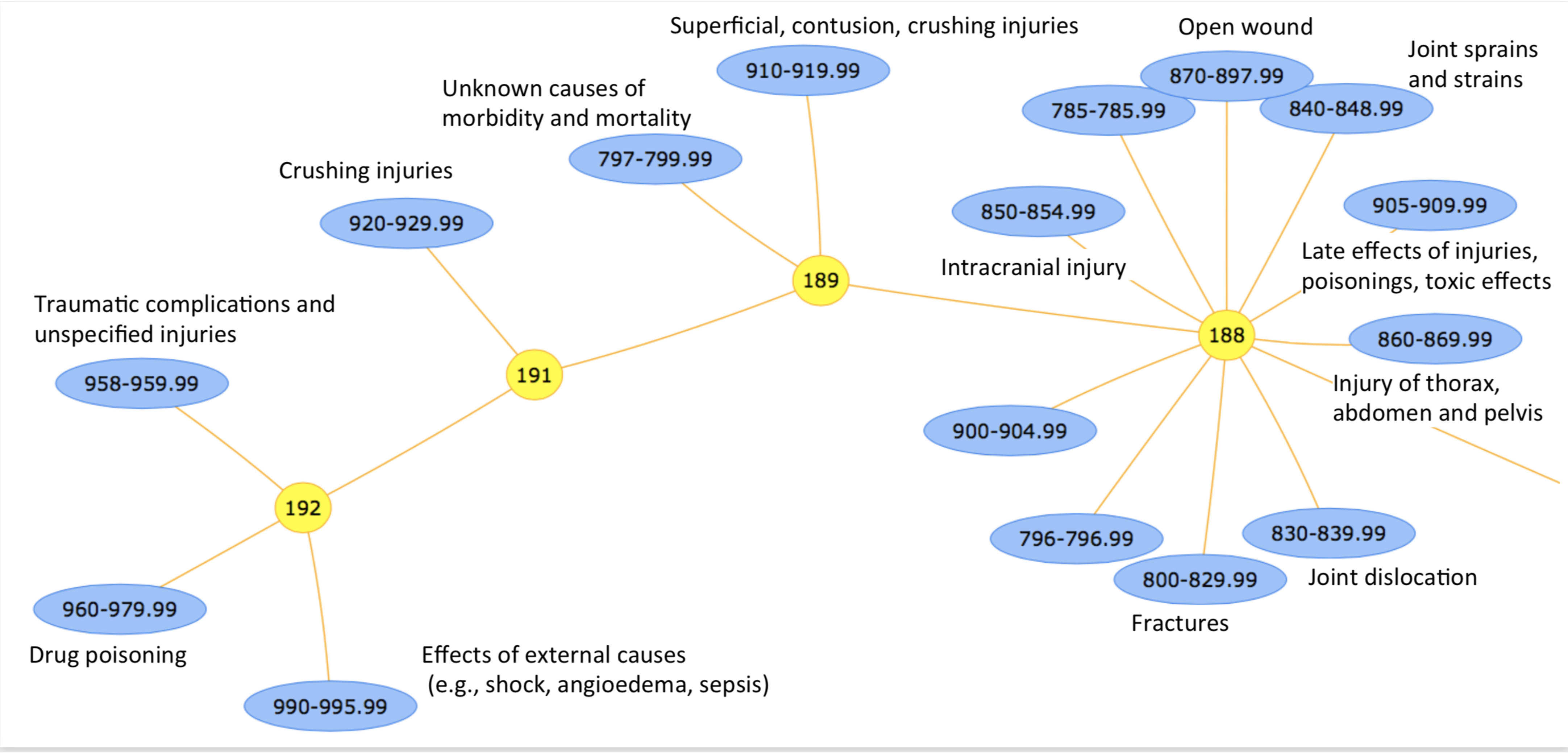}
\end{center}
\caption{ An example of four subtrees which represent groups of similar diseases which may commonly co-occur. Most variables in this subtree are related to trauma.} \label{Fig:tree_mimic2_2}
\end{figure}

\begin{figure}
	\tikzstyle{every node}=[font=\scriptsize]
	\begin{minipage}[t]{0.3\textwidth}
% Minimal pgfplots version: 1.3
%  http://www.mathworks.com/matlabcentral/fileexchange/22022-matlab2tikz
\begin{tikzpicture}

\begin{axis}[%
width=1.42638in,
height=1.125in,
at={(0in,0in)},
scale only axis,
separate axis lines,
every outer x axis line/.append style={black},
every x tick label/.append style={font=\color{black}},
xmin=0,
xmax=1800000,
xlabel={Number of samples},
every outer y axis line/.append style={black},
every y tick label/.append style={font=\color{black}},
ymin=0,
ymax=9000,
ylabel={Running time (seconds)},
title={(a) Running time vs Number of samples},
yticklabel style={font=\footnotesize}
]
\addplot [color=red,solid,mark=asterisk,mark options={solid},forget plot]
  table[row sep=crcr]{%
410964	3676\\
821928	4524\\
1232892	8148\\
1643857	8965\\
};
\end{axis}
\end{tikzpicture}%
	\end{minipage}%
	\hspace{0.5cm}
	\begin{minipage}[t]{0.3\textwidth}
% Minimal pgfplots version: 1.3
%  http://www.mathworks.com/matlabcentral/fileexchange/22022-matlab2tikz
\begin{tikzpicture}

\begin{axis}[%
width=1.42638in,
height=1.125in,
at={(0in,0in)},
scale only axis,
separate axis lines,
every outer x axis line/.append style={black},
every x tick label/.append style={font=\color{black}},
xmin=0,
xmax=1000,
xlabel={Number of observed nodes},
every outer y axis line/.append style={black},
every y tick label/.append style={font=\color{black}},
ymin=0,
ymax=18000,
ylabel={Running time (seconds)},
title={(b) Running time vs Number of nodes},
yticklabel style={font=\footnotesize}
]
\addplot [color=red,solid,mark=asterisk,mark options={solid},forget plot]
  table[row sep=crcr]{%
238	2491\\
476	6792\\
714	11272\\
952	16291\\
};
\end{axis}
\end{tikzpicture}%
	\end{minipage}%
	\hspace{0.5cm}
	\begin{minipage}[t]{0.3\textwidth}
% Minimal pgfplots version: 1.3
%  http://www.mathworks.com/matlabcentral/fileexchange/22022-matlab2tikz
\begin{tikzpicture}

\begin{axis}[%
width=1.42638in,
height=1.125in,
at={(0in,0in)},
scale only axis,
separate axis lines,
every outer x axis line/.append style={black},
every x tick label/.append style={font=\color{black}},
xmin=0,
xmax=70,
xlabel={Number of threads},
every outer y axis line/.append style={black},
every y tick label/.append style={font=\color{black}},
ymin=0,
ymax=70,
ylabel={Speed-up factor},
title={(c) Speed-up vs available threads},
legend style={at={(0.03,0.97)},anchor=north west,legend cell align=left,align=left,fill=none,draw=none},
yticklabel style={font=\footnotesize}
]
\addplot [color=red,solid,mark=asterisk,mark options={solid}]
  table[row sep=crcr]{%
1	1\\
8	7.9\\
16	14.7565314862185\\
32	26.4474321331067\\
64	45.2037330515936\\
};
\addlegendentry{Method speed-up};

\addplot [color=blue,dotted,mark=o,mark options={solid}]
  table[row sep=crcr]{%
1	1\\
8	8\\
16	16\\
32	32\\
64	64\\
};
\addlegendentry{Ideal speed-up};

\end{axis}
\end{tikzpicture}%
	\end{minipage}%
	\vspace{-0.8em}
	\caption{\textbf{(a)} CMS dataset sub-sampling w.r.t.~varying number
		of samples. \textbf{(b)} MIMIC2 dataset sub-sampling w.r.t.~varying number of observed nodes.
		Each one of the observed nodes is binary ($d = 2$).
		\textbf{(c)} MIMIC2 dataset: Scaling w.r.t. varying computational power, establishing the
		scalability of our method even in the large $p$ regime.
		The number of observed nodes is $1083$ and each
		one of them is binary ($p = 1083, d = 2$).}
	\label{Fig: scale}
\end{figure}

\section{ADDITIVITY OF THE MULTIVARIATE INFORMATION DISTANCE}
\label{apdx:additive}
Recall that the additive information distance between nodes two categorical variables $x_i$ and $x_j$ was defined in \cite{choi2011learning}. We extend the notation of information distance to high dimensional variables via Definition~\ref{def:info_dist} and present the proof of its additivity in Lemma~\ref{lem:additive} here.
\begin{proof}
\[
\mathbb{E}[x_a x_c^\top]= \mathbb{E}[\mathbb{E}[x_a x_c^\top | x_b]] = A \mathbb{E}[x_b x_b^\top] B^\top
\]
%Here $a -> b -> c$ is the graph, and $A = P(x_a | x_b)$ and $B=P(x_c|x_b)$.
Consider three nodes $a, b, c$ such that there are edges between $a$ and $b$, and $b$ and $c$. Let the $A = \mathbb{E}(x_a | x_b)$ and $B=\mathbb{E}(x_c|x_b)$. From Definition~\ref{def:info_dist}, we have, assuming that $\mathbb{E}(x_a x_a^\top)$, $\mathbb{E}(x_b x_b^\top)$ and $\mathbb{E}(x_c x_c^\top)$ are full rank.
%fhcomment{Is this true in our yelp case? If this is true, we are in trouble......}
\begin{align*}
\dist(v_a,v_c) & = -\log \frac{  \prod\limits_{i=1}^{k}\sigma_i(\mathbb{E}(x_a x_c^\top))    }{    \sqrt{ \det (\mathbb{E}(x_a x_a^\top ) )  \det( \mathbb{E}(x_cx_c^\top) )}  }\\
e^{-\dist(v_a,v_c)} & = \det\left(  \mathbb{E}(x_ax_a^\top)^{-1/2} U^\top \mathbb{E}(x_a x_c^\top)  V \mathbb{E}(x_c x_c^\top)^{-1/2}\right)
\end{align*}
%where $k-svd(\mathbb{E}(x_ax_c^\top))= U \Sigma V^\top$.
where $k$-SVD$((\mathbb{E}(x_ax_c^\top))= U \Sigma V^\top)$.
Similarly,
\begin{align*}
e^{-\dist(v_a,v_b)} & = \det\left(  \mathbb{E}(x_ax_a^\top)^{-1/2} U^\top \mathbb{E}(x_a x_b^\top)  W \mathbb{E}(x_b x_b^\top)^{-1/2}\right)\\
e^{-\dist(v_b,v_c)} & = \det\left(  \mathbb{E}(x_bx_b^\top)^{-1/2} W^\top \mathbb{E}(x_b x_c^\top)  V \mathbb{E}(x_c x_c^\top)^{-1/2}\right)
\end{align*} where $k$-SVD$((\mathbb{E}(x_ax_b^\top))= U \Sigma W^\top)$ and $k$-SVD$((\mathbb{E}(x_bx_c^\top))= W \Sigma V^\top)$.

Therefore,
\begin{align*}
e^{-(\dist(a,b)+\dist(b,c))} & = \det(  \mathbb{E}(x_ax_a^\top)^{-1/2}  U^\top \mathbb{E}(x_a x_b^\top)   \mathbb{E}(x_b x_b^\top)^{-1/2 -1/2} \mathbb{E}(x_b x_c^\top) V   \mathbb{E}(x_c x_c^\top)^{-1/2}     )\\
& = \det (\mathbb{E}(x_a x_a^\top)^{-1/2}  U^\top A \mathbb{E}(x_b x_b^\top) B^\top V \mathbb{E}(x_c x_c^\top)^{-1/2} ) = e^{-\dist(v_a,v_c)} 
\end{align*}
%Note that $\mathbb{E}(x_a x_b^\top)= A \mathbb{E}(x_b x_b^\top)$.
We conclude that the multivariate information distance is additive. Note that $\mathbb{E}\left[x_a x_b^\top\right] = \mathbb{E}\left(\mathbb{E}\left( x_a x_b^\top \lvert x_b\right)\right)=\mathbb{E}\left(A x_b x_b^\top\right)= A \mathbb{E}(x_b x_b^\top)$.
%\fhcomment{need to add the discuss when  $\mathbb{E}(x_a x_a^\top)$ is not full rank.}
\end{proof}
We note that when the second moments are not full rank, the above distance can be extended as follows:
\[
\dist(v_a,v_c) = -\log \frac{  \prod\limits_{i=1}^{k}\sigma_i(\mathbb{E}(x_a x_c^\top))    }{    \sqrt{  \prod\limits_{i=1}^{k}\sigma_i(\mathbb{E}(x_a x_a^\top))    \prod\limits_{i=1}^{k}\sigma_i(\mathbb{E}(x_c x_c^\top))     }  }.
\]

%%%%%%%%%%%%%%%%%%%%%
\section{LOCAL RECURSIVE GROUPING}
\label{apdx:plrg}
The Local Recursive Grouping (LRG) algorithm is a local divide and conquer procedure for learning the structure and parameter of the latent tree (Algorithm~\ref{algo:plrg}).
We perform recursive grouping simultaneously on the sub-trees of the MST. 
Each of the sub-tree consists of an internal node and its neighborhood nodes. 
We keep track of the internal nodes of the MST, and their neighbors. 
The resultant latent sub-trees after LRG can be merged easily to recover the final latent tree. 
Consider a pair of neighboring sub-trees in the MST. 
They have two common nodes (the internal nodes) which are neighbors on MST. 
Firstly we identify the path from one internal node to the other in the trees to be merged, then compute the multivariate information distances between the internal nodes and the introduced hidden nodes. 
We recover the path between the two internal nodes in the merged tree by inserting the hidden nodes closely to their surrogate node. Secondly, we merge all the leaves which are not in this path by attaching them to their parent. 
Hence, the recursive grouping can be done in parallel and we can recover the latent tree structure via this merging method.

\begin{lemma}\label{lem:surrogate}
If an observable node $v_j$ is the surrogate node of a hidden node $h_i$, then the hidden node $h_i$ can be discovered using $v_j$ and the neighbors of $v_j$ in the MST.
\end{lemma}  
This is due to the additive property of the multivariate information distance on the tree and the definition of a surrogate node. 
This observation is crucial for a completely local and parallel structure and parameter estimation.  It is also easy to see that all internal nodes in the MST are surrogate nodes. 

After the parallel construction of the MST, we look at all the internal nodes $\mathcal{X}_\text{int}$. For $v_i \in \mathcal{X}_\text{int}$, we denote the neighborhood of $v_i$ on MST as $\Nb\text{sub}(v_i;\text{MST})$ which is a small sub-tree. Note that the number of such sub-trees is equal to the number of internal nodes in MST.

For any pair of sub-trees, $\Nb_{\text{sub}}(v_i;\text{MST})$ and $\Nb_{\text{sub}}(v_j;\text{MST})$, there are two topological relationships, namely overlapping (i.e., when the sub-trees share at least one node in common) and non-overlapping (i.e., when the sub-trees do not share any nodes).

Since we define a neighborhood centered at $v_i$ as only its immediate neighbors and itself on MST, the overlapping neighborhood pair $\Nb_{\text{sub}}(v_i;\text{MST})$ and $\Nb_{\text{sub}}(v_j;\text{MST})$ can only have conflicting paths, namely path$(v_i,v_j; \Adj_i)$ and path$(v_i,v_j;\Adj_j)$, if $v_i$ and $v_j$ are neighbors in MST.

With this in mind, we locally estimate all the latent sub-trees, denoted as $\Adj_i$, by applying Recursive Grouping~\citep{choi2011learning} in a parallel manner on $\Nb\text{sub}(v_i;\text{MST}) ,\ \forall v_i \in \mathcal{X}_\text{int}$. Note that the latent nodes automatically introduced by $\RG(v_i)$ have $v_i$ as their surrogate.  We update the tree structure by joining each level in a bottom-up manner. The testing of the relationship among nodes~\citep{choi2011learning} uses the additive multivariate information distance metric (Appendix~\ref{apdx:additive}) $\Phi(v_i,v_j;k)=\dist(v_i,v_k)-\dist(v_i,v_k)$ to decide whether the nodes $v_i$ and $v_j$ are parent-child or siblings. If they are siblings, they should be joined by a hidden parent. If they are parent and child, the child node is placed as a lower level node and we add the other node as the single parent node, which is then joined in the next level.

Finally, for each internal edge of MST connecting two internal nodes $v_i$ and $v_j$, we consider merging the latent sub-trees. In the example of two local estimated latent sub-trees in Figure~\ref{Fig:StructureLearning}, we illustrate the complete local merging algorithm that we propose.  

%\begin{figure}[h]
%  \centering
%\includegraphics[width=\textwidth]{../figures/StructureLearning.eps}
%  \caption{\small \textbf{(a)} Ground truth latent tree, numbers on edges are \emph{multivariate information distances}. \textbf{(b)} MST constructed using the \emph{multivariate information distances}. $v_3$ and $v_5$ are internal (surrogate) nodes. Note that \emph{multivariate information distances} are additive on latent tree, not on MST. \textbf{(c1)} LCR on $\Nb[v_3,\text{MST}]$ to get local structure $\Adj_3$. Pink shadow denotes the active set. Local parameter estimation is carried out over triplets with joint node, such as ($v_2$, $v_3$, $v_5$) with joint node $h_1$. \textbf{(c2)} LCR on $\Nb[v_5,\text{MST}]$ to get local structure $\Adj_5$. Cyan shadow denotes the active set. \textbf{(d1)}\textbf{(d2)} Merging local sub-trees. Path($v_3$,$v_5$; $\Adj_3$) and path($v_3$,$v_5$; $\Adj_5$) conflict.  \textbf{(e)} Final recovery. }
%  \label{Fig:StructureLearning2}
%\end{figure}

\section{PROOF SKETCH FOR THEOREM~\ref{theorem:main}}\label{appen:guarantee}
We argue for the correctness of the method under exact moments. The sample complexity follows from the previous works.
In order to clarify the proof ideas, we define the notion of \emph{surrogate node}~\citep{choi2011learning} as follows.
 \begin{definition}
 \label{def:surrogate}
 Surrogate node for hidden node $h_i$ on the latent tree $\mathcal{T}=(\mathcal{V},\mathcal{E})$ is defined as
 $\text{Sg}(h_i;\mathcal{T}):= \arg \min\limits_{v_j\in \mathcal{X}} \dist(v_i,v_j)$.
 \end{definition}

 In other words, the surrogate for a hidden node is an observable node which has the minimum multivariate information distance from the hidden node. See Figure~\ref{Fig:StructureLearning}(a), the surrogate node of $h_1$, $\text{Sg}(h_1;\mathcal{T})$, is $v_3$, $\text{Sg}(h_2;\mathcal{T}) = \text{Sg}(h_3;\mathcal{T}) = v_5$. Note that the notion of the  surrogate node is only required for analysis, and our algorithm does not need to know this information.  
 
The notion of surrogacy allows us to relate the constructed MST (over observed nodes) with the underlying latent tree. It can be easily shown that contracting the hidden nodes to their surrogates on latent tree leads to MST. Local recursive grouping procedure can be viewed as reversing these contractions, and hence, we obtain consistent local sub-trees. 

We now argue the correctness of the structure union procedure, which merges the local sub-trees. In each reconstructed sub-tree $\Adj_i$, where $v_i$ is the group leader, the discovered hidden nodes $\{h^i\}$ form a  surrogate relationship with $v_i$, i.e. $\Sg(h^i;\calT) = v_i$. Our merging approach maintains these surrogate relationships. For example in Figure~\ref{Fig:StructureLearning}(d1,d2), we have the path $v_3-h_1-v_5$ in $\Adj_3$ and path $v_3-h_3-h_2-v_5$ in $\Adj_5$.
The resulting path is $v_3-h_1-h_3-h_2-v_5$, as seen in Figure~\ref{Fig:StructureLearning}(e). We now argue why this is correct. As discussed before, $\Sg(h_1;\calT)=v_3$ and $\Sg(h_2;\calT)=\Sg(h_3;\calT)=v_5$. When we merge the two subtrees, we want to preserve the paths from the group leaders to the added hidden nodes, and this ensures that the surrogate relationships are preserved in the resulting merged tree. Thus, we obtain a global consistent tree structure by merging the local structures. The correctness of parameter learning comes from the consistency of the tensor decomposition techniques and careful alignments of the hidden labels across different decompositions.
Refer to Appendix~\ref{appen:correctness},~\ref{appen:samplecomp} for proof details and the sample complexity.

 \section{PROOF OF CORRECTNESS FOR LRG}\label{appen:correctness}

\begin{definition}
A latent tree $\mathcal{T}_{\ge3}$ is defined to be a minimal (or identifiable) latent tree if it satisfies that each latent variable has at least 3 neighbors.
\end{definition}

\begin{definition}
Surrogate node for hidden node $h_i$ in  latent tree $\mathcal{T}=(\mathcal{V},\mathcal{E})$ is defined as
\[
\Sg(h_i;\mathcal{T}):= \arg \min\limits_{v_j\in \mathcal{X}} \dist(v_i,v_j).
\]
\end{definition}

There are some useful observations about the MST in~\cite{choi2011learning} which we recall here. 
\begin{property}[MST $-$ surrogate neighborhood preservation]\label{prop:MST1}
The surrogate nodes of any two neighboring nodes in $\mathcal{E}$ are also neighbors in the MST. I.e.,
\[
(h_i,h_j)\in \mathcal{E} \Rightarrow (\text{Sg}(h_i),\text{Sg}(h_j)) \in \text{MST}.
\]
\end{property}

\begin{property}[MST $-$ surrogate consistency along path]\label{prop:MST2}
If $v_j\in\mathcal{X}$ and $v_h\in \text{Sg}^{-1}(v_j)$, then every node along the path connecting $v_j$ and $v_h$ belongs to the inverse surrogate set $\text{Sg}^{-1}(v_j)$, i.e.,
\[
v_i \in \text{Sg}^{-1}(v_j) ,\ \forall v_i\in \text{Path}(v_j,v_h)
\]
if 
\[
v_h\in \text{Sg}^{-1}(v_j).
\]
\end{property}
 
 The MST properties observed connect the MST over observable nodes with the original latent tree $\mathcal{T}$.  We obtain MST by contracting all the latent nodes to its surrogate node.

 Given that the correctness of CLRG algorithm is proved in~\cite{choi2011learning}, we prove the equivalence between the CLRG and PLRG.
 
 \begin{lemma}
 For any sub-tree pairs $\Nb[v_i;\text{MST}]$ and $\Nb[v_i;\text{MST}]$, there is at most one overlapping edge. The overlapping edge exists if and only if $v_i \in \Nb(v_j;\text{MST})$.
 \end{lemma}
 
 This is easy to see. 
 
 \begin{lemma}
 Denote the latent tree recovered from $\Nb[v_i;\text{MST}]$ as $\Adj_i$ and similarly for $\Nb[v_j;\text{MST}]$.  
 The inconsistency, if any, between $\Adj_i$  and  $\Adj_j$ occurs in the overlapping path$(v_i,v_j;\Adj_i)$ in and  path$(v_i,v_j;\Adj_j)$ after LRG implementation on each subtrees. 
 \end{lemma}
 
 We now prove the correctness of LRG. Let us denote the latent tree resulting from merging a subset of small latent trees as $T_{\text{LRG}}(S)$, where $S$ is the set of center of subtrees that are merged pair-wisely. CLRG algorithm in~\cite{choi2011learning} implements the RG in a serial manner. Let us denote the latent tree learned at iteration $i$ from CLRG is $T_{\text{CLRG}}(S)$, where $S$ is the set of internal nodes visited by CLRG at current iteration . We prove the correctness of LRG by induction on the iterations.
 
 At the initial step $S = \emptyset$:  $T_{\text{CLRG}}= MST$ and $T_{\text{LRG}}=MST$, thus $T_{\text{CLRG}}=T_{\text{LRG}}$.
 
 Now we assume that for the same set $S_{i-1}$, $T_{\text{CLRG}}=T_{\text{LRG}}$ is true for $r=1,\ldots,i-1$. 
 At iteration $r=i$ where CLRG employs RG on the immediate neighborhood of node $v_i$ on $T_{\text{CLRG}}(S_{i-1})$, let us assume that $H_i$ is the set of hidden nodes who are immediate neighbors of $i-1$.   The CLRG algorithm thus considers all the neighbors and implements the RG. We know that the surrogate nodes of every latent node in $H_i$ belong to previously visited nodes $S_{i-1}$. According to Property~\ref{prop:MST1} and~\ref{prop:MST2}, if we contract all the hidden node neighbors to their surrogate nodes, CLRG thus is a RG on neighborhood of $i$ on MST.  
 
 As for our LRG algorithm at this step, $T_{\text{LRG}}(S_i)$ is the merging between $T_{\text{LRG}}(S_{i-1})$and $\Adj_i$.   %(Note that the merging is parallel, but in order to prove the correctness, we only look at parts of the merging. )
 The latent nodes whose surrogate node is $j$ are introduced between the edge $(i-1,i)$.   Now that we know $\Adj_i$ is the RG output from immediate neighborhood of $i$ on MST.
 Therefore, we proved that $T_{\text{CLRG}}(S_i)= T_{\text{LRG}}(S_i)$.

\section{CROSS GROUP ALIGNMENT CORRECTION}\label{appen:alignment}
In order to achieve cross group alignments,  tensor decompositions on two cross group triplets have to be computed. The first triplet is formed by three nodes: reference node in group 1, $x_1$, non-reference node in group 1, $x_2$, and reference node in group 2, $x_3$.  The second triplet is formed by three nodes as well: reference node in group 2, $x_3$, non-reference node in group 2, $x_4$ and reference node in group 1, $x_1$. Let us use $h_1$ to denote the parent node in group 1, and $h_2$ the parent node in group 2. 
 
 From $\Trip(x_1,x_2,x_3)$%glob_ref, ref_mem, local_ref>$ , we know
, we obtain $P(h_1|x_1) = \tilde{A}$,
%\[
%P(x_1| h_1) ==>P(h_1|x_1) = \tilde{A}
%\]
$P(x_2|h_1) = B$ and 
$P(x_3|h_1) = P(x_3|h_2) P(h_2|h_1) = DE$.
From $\Trip(x_3,x_4,x_1)$, we know
%<local_ref,local_mem,  glob_ref>$, we know
$P(x_3|h_2) = D\Pi$, $P(x_4|h_2) = C\Pi$ and $P(h_2|x_1) = P(h_2|h_1)P(h_1|x_1) = \Pi E\tilde{A}$, where $\Pi$ is a permutation matrix.
%\[P(x_1|h_2)==>P(h_2|x_1)=P(h_2|h_1)P(h_1|x_1) = \Pi E\tilde{A}.\]
We compute $\Pi$ as $\Pi = \sqrt{ (\Pi E\tilde{A})  (\tilde{A})^{\dag}  (DE)^{\dag} (D\Pi)}$ so that $D =(D\Pi) \Pi^{\dag}$ is aligned with group 1. Thus, when all the parameters in the two groups are aligned by permute group 2 parameters using $\Pi$, thus the alignment is completed.%, these two families are aligned.

 Similarly, the alignment correction can be done by calculating the permutation matrices while merging different threads.
 
 Overall, we merge the local structures and align the parameters from LRG locla sub-trees using Procedure~\ref{algo:pmac} and~\ref{algo:alignment}.

\section{COMPUTATIONAL COMPLEXITY}\label{appen:compuComplex}
We recall some notations here: $d$ is the observable node dimension, $k$ is the hidden node dimension ($k \ll d$), $N$ is the number of samples, $p$ is the number of observable nodes, and $z$ is the number of non-zero elements in each sample.

Multivariate information distance estimation involves sparse matrix multiplications to compute the pairwise second moments. Each observable node has a $d \times N$ sample matrix with $z$ non-zeros per column. Computing the product $x_1 x_2^T$ from a single sample for nodes $1$ and $2$ requires $O(z)$ time and there are $N$ such sample pair products leading to $O(Nz)$ time. There are $O(p^2)$ node pairs and hence the degree of parallelism is $O(p^2)$. Next, we perform the $k$-rank SVD of each of these matrices. Each SVD takes $O(d^2 k)$ time using classical methods. Using randomized methods~\citep{gittens2013revisiting}, this can be improved to $O(d+k^3)$. %We stress that usually either $p$ is large or $d$ is large, not both. 

Next on, we construct the MST in $O(\log p)$ time per worker with $p^2$ workers. The structure learning can be done in $O(\Gamma^3)$ per sub-tree and the local neighborhood of each node can be processed completely in parallel. We assume that the group sizes $\Gamma$ are constant (the sizes are determined by the degree of nodes in the latent tree and homogeneity of parameters across different edges of the tree. The parameter estimation of each triplet of nodes consists of implicit stochastic updates involving products of $k \times k$ and $d \times k$ matrices. Note that we do not need to consider all possible triplets in groups but each node must be take care by a triplet and hence there are $O(p)$ triplets. This leads to a factor of $O(\Gamma k^3 + \Gamma d k^2)$ time per worker with $p/\Gamma$ degree of parallelism.

At last, the merging step consists of products of $k \times k$ and $d \times k$ matrices for each edge in the latent tree leading to $O(d k^2)$ time per worker with $p/\Gamma$ degree of parallelism.

\section{SAMPLE COMPLEXITY}\label{appen:samplecomp}

%We consider the sample complexity of our proposed parallel latent tree algorithm in two phases, namely the number of samples required for the faithful structure recovery and number of samples for the parameter estimation. According to~\cite{anandkumar2011spectral} and~\cite{anandkumar2012two} the number of samples required are discussed in Appendix~\ref{appen:samplecomp}.

From Theorem 11 of~\cite{choi2011learning}, we recall the number of samples required for the recovery of the tree structure that is consistent with the ground truth (for a precise definition of consistency, refer to Definition 2 of~\cite{choi2011learning}).%faithful recovery of the latent tree structure.

%\begin{lemma}
%If
%\begin{equation}
%N > \frac{    200 k^2 B^2 t       }       {   \left(  \frac{ \gamma_{\min}^2	}  {\gamma_{\max}}       (1-\dist_{\max}) \right)^2} + \frac{7 kM^2 t} {           \frac{ \gamma_{\min}^2	}  {\gamma_{\max}}     (1-\dist_{\max})     },
%%N > \frac{}{\frac{\gamma_{\text{min}}^2}{\gamma_{\text{max}}(1-\rou_{\text{max}})}}
%\end{equation}
%then with probability at least $1-\eta$, proposed algorithm returns $\widehat{\mathcal{T}} =\mathcal{T}$, where
%$$B:=\max_{x_i,x_j\in \mathcal{X}} \left\{           \sqrt{  \max\{ \lVert \mathbb{E} [\lVert x_i\rVert^2 x_jx_j^\top]\rVert \},  \max\{ \lVert \mathbb{E} [\lVert x_j\rVert^2 x_ix_i^\top]\rVert \}          }        \right\},$$
%$$M:= \max_{x_i\in \mathcal{X}} \left\{     \lVert x_i\rVert   \right\},$$
%$$t:= \max_{x_i,x_j\in \mathcal{X}} \left\{              	4\ln(4                       	\frac{\mathbb{E}[\lVert x_i\rVert^2 \lVert x_j\rVert^2] -\Tr(\mathbb{E}[x_ix_j^\top]\mathbb{E}[x_jx_i^\top])}{\max\{  \lVert \mathbb{E}[\lVert x_j\rVert^2x_ix_i^\top]\rVert, \lVert \mathbb{E}[\lVert x_i\rVert^2x_jx_j^\top]\rVert \}}			  n/\eta)					   \right\}.$$
%$$\gamma_{\min}:= \min\limits_{\{x_1,x_2\}} {\{\sigma\left(\mathbb{E}[x_1x_2^\top] \right)\}}
%$$
%$$\gamma_{\max}:= \max\limits_{\{x_1,x_2\}} {\{\sigma\left(\mathbb{E}[x_1x_2^\top] \right)\}}
%$$
%\end{lemma}

From~\cite{anandkumar2012two}, we recall the sample complexity for the faithful recovery of parameters via tensor decomposition methods.

We define $\epsilon_P$ to be the noise raised between empirical estimation of the second order moments and exact second order moments, and $\epsilon_T$ to be the noise raised between empirical estimation of the third order moments and the exact third order moments.
\begin{lemma}\label{lm:parameter_robustness}
Consider positive constants $C$, $C'$, $c$ and $c'$, the following holds. If
\begin{align*}
 \epsilon_P &\le c \frac{\frac{\lambda_k}{\lambda_1}}{k}, \quad \quad \epsilon_T\le c' \frac{\lambda_k \sigma_k^{3/2}}{k}\\
 N &\ge C\left(\log(k) + \log\left(\log\left(	\frac{\lambda_1 \sigma_k^{3/2}}{\epsilon_T}   +\frac{1}{\epsilon_P}		\right)\right)\right)\\
 L&\ge \poly(k)\log(1/\delta),
 \end{align*} then with probability at least $1-\delta$, tensor decomposition returns $(\widehat{v_i},\lambda_i): i\in[k]$ satisfying, after appropriate reordering,

 \begin{align*}
 \lVert \widehat{v_i} - v_i\rVert_2 &\le C' \left( \frac{1}{\lambda_i} \frac{1}{\sigma_k^2} \epsilon^T  + \left(\frac{\lambda_1}{\lambda_i}\frac{1}{\sqrt{\sigma_k}}+1\right)\epsilon_P		\right)\\
 \lvert \widehat{\lambda_i} - \lambda_i\rvert &\le C' \left(\frac{1}{\sigma_k^{3/2}}\epsilon_T+\lambda_1\epsilon_P\right)
 \end{align*}
for all $i\in[k]$.
\end{lemma}
We note that $\sigma_1\ge \sigma_2\ge \ldots \sigma_k >0$ are the non-zero singular values of the second order moments, %$\mathbf{P}$. 
$\lambda_1\ge\lambda_2\ge \ldots \ge \lambda_k>0$ are the ground-truth eigenvalues of the third order moments, and $v_i$ are the corresponding eigenvectors for all $i\in[k]$.

\section{EFFICIENT  SVD USING SPARSITY AND DIMENSIONALITY REDUCTION}\label{apdx:svd}
Without loss of generality, we assume that a matrix whose SVD we aim to compute has no row or column which is fully zeros, since, if it does have zero entries, such row and columns can be dropped.

Let $A \in \mathbb{R}^{n\times n}$ be the  matrix to do SVD. Let $\Phi \in R^{d \times \tilde{k}}$, where $\tilde{k} = \alpha k$ with $\alpha$ is a scalar, usually, in the range $[2,3]$. For the $i^{th}$ row of $\Phi$, if $\sum_i |\Phi|(i,:)\neq 0$ and $\sum_i |\Phi|(:,i)\neq 0$, then there is only one non-zero entry and that entry is uniformly chosen from $[\tilde{k}]$. If either $\sum_i |\Phi|(i,:)= 0$ or $\sum_i |\Phi|(:,i)=0$, we leave that row blank. Let $D\in \mathbb{R}^{d \times d}$ be a diagonal matrix with iid Rademacher entries, i.e., each non-zero entry is $1$ or $-1$ with probability $\frac{1}{2}$. Now, our embedding matrix~\citep{clarkson2013low} is $S = D \Phi $, i.e., we find $AS$ and then proceed with the Nystrom~\citep{DBLP:journals/corr/HuangNHVA13} method. Unlike the usual Nystrom method~\citep{DBLP:journals/corr/abs-1303-1849} which uses a random matrix for computing the embedding, we improve upon this by using a sparse matrix for the embedding since the sparsity improves the running time and the memory requirements of the algorithm.

\end{document}